\documentclass{article}

\usepackage[main,nonatbib,preprint]{neurips_2026}
\usepackage[numbers,sort&compress]{natbib}

\usepackage[utf8]{inputenc} 
\usepackage[T1]{fontenc}    
\usepackage{url}            
\usepackage{booktabs}       
\usepackage{amsfonts}       
\usepackage{amsmath}
\usepackage{amssymb} 
\usepackage{amsthm}
\usepackage{stmaryrd}
\usepackage{nicefrac}       
\usepackage{microtype}      
\usepackage[dvipsnames,table]{xcolor}         
\usepackage{algorithm}
\usepackage{algpseudocode}
\usepackage{caption}
\usepackage{listings}
\usepackage{mathtools}
\usepackage{multirow}
\usepackage{multicol}
\usepackage{tcolorbox}
\usepackage{dsfont}
\usepackage{xspace}
\usepackage{blkarray}
\usepackage{pgfplots}
\usepackage{tabularx}
\usepackage{arydshln}
\pgfplotsset{compat=1.18}
\usepackage{pgfplotstable}
\usepgfplotslibrary{groupplots}
\usepackage[inline]{enumitem}
\usepackage{colortbl}
\newcommand{\gcell}[2]{\cellcolor{ForestGreen!#1}#2}
\newcommand{\rcell}[2]{\cellcolor{BrickRed!#1}#2}

\lstdefinestyle{pyinline}{
  language=Python,
  basicstyle=\ttfamily\scriptsize,
  keywordstyle=\color{MidnightBlue},
  commentstyle=\color{Gray},
  stringstyle=\color{ForestGreen},
  showstringspaces=false,
  columns=fullflexible,
  keepspaces=true,
  breaklines=true,
  frame=single,
  framerule=0.3pt,
  rulecolor=\color{Gray!35},
  backgroundcolor=\color{Gray!5},
  xleftmargin=0.5em,
  xrightmargin=0.5em,
  aboveskip=0.5em,
  belowskip=0.5em
}

\usepackage{tikz}
\usetikzlibrary{automata, positioning, arrows.meta, shapes.geometric}
\usepackage[capitalize,noabbrev]{cleveref}

\definecolor{FilterBlue}{HTML}{2F6FAE}
\definecolor{GainGreen}{HTML}{2E8B57}
\definecolor{DropRed}{HTML}{B44A3A}
\definecolor{SoftGray}{HTML}{F3F4F6}

\definecolor{SoftBlack}{HTML}{6B7280}
\definecolor{SoftGrayLine}{HTML}{9CA3AF}
\definecolor{SoftBlue}{HTML}{5B8DB8}
\definecolor{SoftGreen}{HTML}{6FAE8D}

\newcommand{\triggerbar}[5]{%
  \begin{minipage}[t]{\linewidth}
    \footnotesize #1 {\scriptsize\textcolor{black!55}{[#2]}}\\[-0.35ex]
    \makebox[32mm][l]{%
      \textcolor{black!15}{\rule{32mm}{0.75ex}}%
      \hspace*{-32mm}%
      \pgfmathsetlengthmacro{\triggerbarwidth}{32mm * min(max(#5,0),100) / 100}%
      \textcolor{FilterBlue}{\rule{\triggerbarwidth}{0.75ex}}%
    }%
    \hspace{0.5em}{\scriptsize (#4 events, #5\%)}%
  \end{minipage}
}

\theoremstyle{plain}
\newtheorem{theorem}{Theorem}[section]

\theoremstyle{definition}
\newtheorem{definition}[theorem]{Definition}

\theoremstyle{remark}

\newtheorem{example}[theorem]{Example}

\makeatletter
\newtheorem*{rep@theorem}{\rep@title}
\newcommand{\newreptheorem}[2]{%
\newenvironment{rep#1}[1]{%
 \def\rep@title{#2 \ref{##1}}%
 \begin{rep@theorem}}%
 {\end{rep@theorem}}}
\makeatother

\newcommand{\invalid}{\mathit{ivld}}

\newcommand{\pfilter}{\phi}

\newreptheorem{theorem}{Theorem}
\newreptheorem{lemma}{Lemma}

\newcommand{\rone}{(\emph{i})~}
\newcommand{\rtwo}{(\emph{ii})~}
\newcommand{\rthree}{(\emph{iii})~}
\newcommand{\rfour}{(\emph{iv})~}

\newcommand{\lbbar}{\{\kern-0.5ex|}
\newcommand{\rbbar}{|\kern-0.5ex\}}

\newcommand{\biglbbar}{\left\{\kern-0.5ex\left|}
\newcommand{\bigrbbar}{\right|\kern-0.5ex\right\}}

\newcommand{\name}{\textsc{Palla}\xspace}

\newcommand{\Etrue}{{{\mathsf{true}}}}
\newcommand{\Efalse}{{{\mathsf{false}}}}

\newcommand{\aL}{\mathcal{L}}

\newcommand{\Omit}[1]{}

\newcounter{resq}

\newcommand{\aO}{\mathcal{O}}

\newcommand{\set}[1]{\{{#1}\}}

\definecolor{dgreen}{RGB}{0,128,0}
\definecolor{dred}{RGB}{200,0,0}

\title{Learning the Error Patterns of Language Models}

\author{%
  Jinwoo Kim, Taylor Berg-KirkPatrick, Loris D'Antoni \\
  Department of Computer Science and Engineering \\
  University of California-San Diego \\
  La Jolla, CA 92037 \\
  \texttt{jik083@ucsd.edu}
}

\begin{document}

\maketitle

\begin{abstract}
  When generating outputs for domains with specific validity constraints 
  (e.g., a program should compile),
  LLMs often fail in a small number of focused ways: for example, 
  by using Python function names when generating TypeScript.
  We observe that these error patterns can be represented 
  using a small number of constraints that can be learned in practice. 
  We propose \emph{prefix filters}, which are per-domain-and-LLM symbolic functions,
  as objects to capture the error patterns, 
  \name as an 
  algorithm to learn prefix filters efficiently in practice, 
  and implement \name. 
    Prefix filters learned by \name 
    \rone help us quantitatively analyze the error patterns of LLMs, and 
    \rtwo can be used to constrain the outputs of a model via constrained sampling algorithms.
    For example, \name boosts compile rates for Qwen2.5-1.5B on TypeScript generation, 
    by over 60\%, allowing Qwen2.5-1.5B to achieve similar performance 
    to Llama3.1-8B unconstrained.
\end{abstract}
\section{Introduction}
\label{sec:introduction}

When using large language models (LLMs) in practice, 
  for many domains there exists a clear set of \emph{bad outputs} that we want the LLM to avoid.
  Modern LLMs are far from perfect and sometimes end up generating these bad outputs:
  for example,
a model summarizing a sensitive document may leak personal information, 
or a model generating code may produce programs that fail to compile.
Techniques such as grammar-constrained decoding~\cite{gcd} help, but are imperfect as the notion 
of bad outputs (e.g., programs that are not type-safe) often cannot be precisely captured via a grammar.

A key observation is that although the space of bad outputs can be complex, 
in practice the errors that LLMs make are not uniformly distributed across this space.
Instead, a given model on a given domain 
tends to fail in a small number 
repetitive ways. 
For example, when generating functions in the MLIR language~\cite{mlir}, 
Qwen2.5-7B~\cite{qwen} frequently produces programs containing the nonexistent operator \texttt{torch.aten.softmax}; 
or, Llama3.1-8B~\cite{llama} often inserts the character \texttt{\%} in invalid locations.
  This observation motivates the central question on controlling model behavior we ask in this paper:

\begin{center}
\textit{
Can we learn and suppress the error patterns that a given LLM actually exhibits on a given domain?
}
\end{center}

\paragraph{Prefix filters.}
This paper answers the above question by proposing \emph{prefix filters}: per-domain-and-LLM
\emph{symbolic} functions (e.g., Python functions)
that analyze the prefix of a sample being generated, 
and decide whether the sample has already gone wrong or is likely to do so in the future.
For example, an MLIR sample prefix containing 
the undefined \texttt{torch.aten.softmax} operator
will fail to 
compile regardless of completion---the goal of prefix filters are to identify such samples early.

Prefix filters are a solution to our central question in two ways: 
First,
they are concrete interpretable functions that capture the error pattern of an LLM.
Second, prefix filters can be used directly in conjunction with constrained sampling algorithms~\cite{cars}, 
(i.e., by early-rejecting prefixes caught by the filters and resampling)
to constrain the output of LLMs towards those that the filters do not catch.

\begin{figure}
\includegraphics[width=\textwidth]{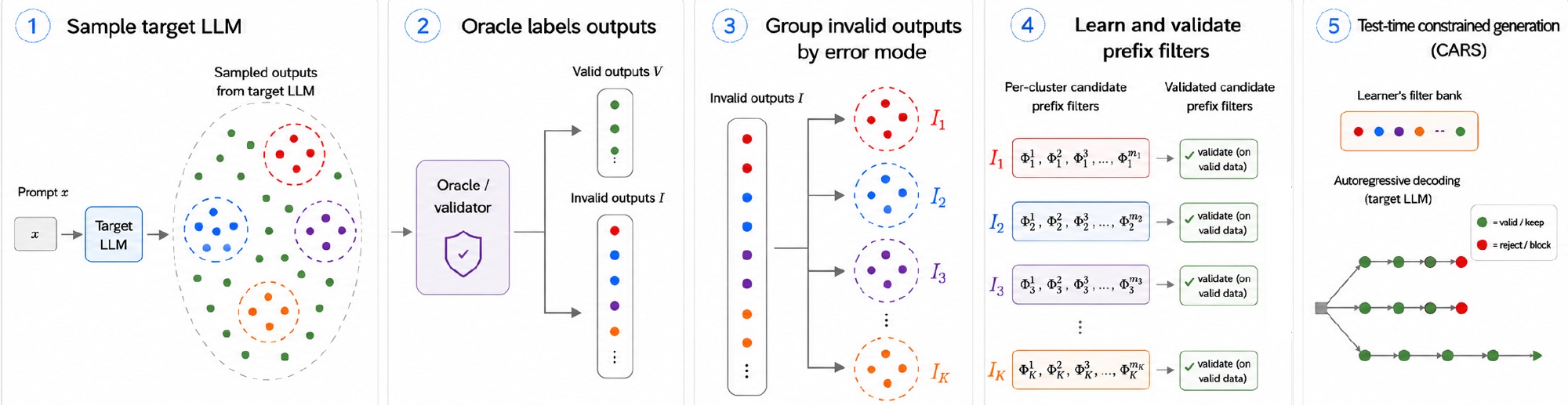}
\caption{Overall pipeline of \name: we exploit the fact that 
error modes are clustered to learn prefix filters 
targeting each cluster, then use the filters to avoid 
samples from the clustered errors at test time.
}
\label{fig:intro-fig}
\end{figure}
Constructing a monolithic object precisely capturing the error pattern of 
an LLM is complex at best, often undecidable~\cite{greatgramma,undec}.
Prefix filters offer a unique solution to this challenge, in that 
we do \emph{not} require that filters be precise. 
Instead, what matters is that the filter is faithful to the 
\emph{actual distribution} of the target LLM; i.e., 
it is OK that the filter excludes some outputs that are valid in principle, as long as these outputs 
are generated \emph{very rarely} by the LLM in question.
This observation allows prefix filters to be simple, efficient, and easy-to-learn, 
all of which are key desiderata.

\paragraph{\name: An algorithm for learning error patterns.}
We propose an algorithm \name for 
learning error patterns from concrete mistakes that LLMs make:
by \rone unconditionally sampling the LLM on domain-specific tasks, 
\rtwo identifying the bad outputs via an external oracle, \rthree grouping the errors, then 
\rfour querying a separate synthesizer on the bad outputs to distill error patterns 
as concrete objects.
\Cref{fig:intro-fig} visualizes \name at a high level.
We use an external LLM as the synthesizer in step 3; 
in principle, one could also use a symbolic program synthesizer~\cite{loud, sygus, semgus} as well.
For example, when generating filters for Llama3.1-8B on the domain of generating TypeScript functions, one might
\rone sample Llama on a variety of specific TypeScript 
tasks (e.g., benchmarks from MultiPL-E~\cite{multiple}); 
\rtwo identify invalid samples using the TypeScript compiler as an oracle, and 
\rthree ask an external LLM to generate patterns from the invalid samples.

\paragraph{Evaluating \name.}
We evaluate prefix filters and \name across
4 domains (generating \rone functions in MLIR, 
\rtwo chemically valid molecules, 
\rthree natural language summaries, 
\rfour TypeScript functions)
and 5 LLMs (Qwen2.5-1.5B/7B/14B~\cite{qwen}, Llama3.1-8B~\cite{llama}, and Gemma3-12B~\cite{gemma}), 
with GPT-5.4-mini~\cite{gptmini} as the synthesizer.
Results show that \name can learn prefix filters from a modest number of unconstrained samples, 
reveal insight on the different error patterns of LLMs,
and is very effective for the task of constrained generation.

\paragraph{Contributions.} 
To summarize, we make the following contributions:

\emph{Prefix filters.} 
We propose the idea of \emph{prefix filters}, per-domain-and-LLM functions that 
recognize a outstanding failure modes of an LLM on the domain in question (\Cref{sec:prefix}).

\emph{Learning prefix filters.}
We propose \name, an algorithm for learning 
error patterns as
prefix filters, without
requiring extensive domain knowledge and human intervention
(\Cref{sec:prefix}).

\emph{Implementation and Evaluation.}
We evaluate the efficacy of \name and prefix filters combined with constrained rejection sampling
on a combination of 4 domains and 5 LLMs, 
showing that \name offers quantitative conclusions about the error patterns of different LLMs 
and is also highly effective for constrained generation (\Cref{sec:experiments} and \ref{sec:results}).
For example, with \name, Qwen-1.5B can achieve similar performance with unconstrained Llama-8B, 
a much larger model, on the task of 
Typescript generation, simply by learning 132 filters that correspond to its error modes.


\section{Errors, Problem Definition, and Related Work}

We now formally define what it means for an LLM to make an `error' on a certain domain, and 
state the \emph{constrained generation problem}, 
which is to constrain the output of the LLM such that it does not make any errors.
We use $\Sigma$ to denote the tokens of an LLM, 
$\Sigma^*$ to denote the set of strings obtainable by concatenating tokens, and 
$s \cdot s'$ for $s, s' \in \Sigma^*$ to denote the concatenation of $s$ and $s'$.

\begin{definition}[Errors and Constrained Generation]
  Let $\aL$ be an LLM $\aL: \Sigma^* \rightarrow \Delta({\Sigma^*})$ 
  that takes as input a string $p$ 
  and outputs a distribution $\Pr_{\aL}(\cdot \mid p)$ of strings.
  Let $D$ be a \emph{domain} and 
  $\aO$ be an \emph{oracle} $\aO: \Sigma^* \rightarrow \set{\Etrue, \Efalse}$ for $D$ that classifies strings $s$
  into good samples ($\aO(s) = \Etrue$) and bad samples ($\aO(s) = \Efalse$).
  Let $O_{\invalid}$ be the set of bad samples $O_{\invalid} = \set{o_{\invalid} \mid \aO(o_{\invalid}) = \Efalse}$.

  Let $P_D$ denote the set of prompts over the domain $D$.
    Given a prompt $p_D \in P_D$, if sampling from $\Pr_{\aL}(\cdot \mid p_D)$ results 
    in an invalid output $o_{\invalid} \in O_{\invalid}$, 
    we say that $\aL$ has made an \emph{error}.
  The \emph{constrained generation problem} for $D$ is to restrict the image of $\aL$
  such that for any $p_D \in P_D$, 
  $\Pr_{\aL}(O_{\invalid} \mid p_D) = 0$
  (i.e., the probability of sampling any invalid string $o_{\invalid} \in O_{\invalid}$ is zero).
  
\end{definition}

  \emph{Rejection sampling}, i.e., re-sampling from $\aL$ until a good sample is drawn, 
  is the most straightforward way to obtain an error-free set of outputs.
  However, rejection sampling is expensive, and is not always feasible online during test time, 
  depending on the oracle.
Existing work thus mainly attempts to solve the constrained generation problem in two ways: 
\rone constrained decoding, and \rtwo fine-tuning.

\paragraph{Constrained decoding}
Constrained decoding
refers to a family of techniques in which one starts with 
a grammar~\cite{gcd,greatgramma,chopchop} 
(or sometimes an automaton~\cite{vechevtypes}) that describes a set of acceptable strings.
Using the automaton, one can then construct token masks that filter out tokens 
which are guaranteed not to have a valid continuation, guaranteeing that 
a generated sample is accepted.

Constrained decoding has many advantages: 
\rone modern implementations are highly efficient~\cite{outlines,llguidance}, 
\rtwo grammars can be reused among different LLMs, and
\rthree it guarantees that produced samples are accepted by the grammar.
However, the main downside of constrained decoding is that grammar acceptance does not precisely align 
with most oracles: properties such as type safety cannot be captured by grammars alone.
As a result, constrained decoding mainly \emph{overapproximates} the oracle and cannot guarantee 
true oracle alignment, resulting in invalid samples under constrained decoding.
Constructing new grammars for domains also remains a complex task that requires human expertise.

\paragraph{Fine-tuning}
Fine-tuning approaches to constrained generation directly update model weights 
so the output distribution better aligns with the target oracle. 
Reinforcement learning-style methods and related objectives~\cite{ouyang2022training,stiennon2020learning,rafailov2023direct,ppo,rlhf,bai2022constitutional, deepseekmath} 
have shown strong results in settings such as summarization and
instruction following, and code-focused variants similarly use execution or 
test feedback to improve program generation~\cite{coderl,supercoder,rlef}. 
These methods are powerful, but typically require 
large amounts of data, compute, and careful design of the reward model.

\name and prefix filters alleviate the problems of existing approaches: 
\name does not require extra compute, data, or human intervention 
to learn prefix filters, and 
tailored per-domain-and-LLM error patterns creates a set of constraints that are much better aligned 
to the target LLM.

\section{Error-Driven Pattern Learning and Prefix Filters}
\label{sec:prefix}
\begin{algorithm}[t]
  \caption{Error-Driven Pattern Learning}
  \label{alg:learning-alg-1}
  \small
  \begin{algorithmic}[1]
  \State \textbf{Input:} 
    LLM $\aL$, oracle $\mathcal{O}$, unconstrained sample budget $B$. \ 
    \textbf{Output:} Set of error patterns $\Phi$

  \State $\Phi \gets \emptyset$
  \State $S \gets \{x_i \sim \aL\}_{i=1}^{B}$ \Comment{Sample outputs from $\aL$ to set $S$}
  \State $V \gets \textsc{ValidSamples}(\aO, S)$ \Comment{Get valid samples from $S$}
  \State $I \gets \textsc{InvalidSamplesByGroup}(\aO, S)$ \Comment{Group invalid samples from $S$ according to cause}
  \State \textbf{for each} $I_e \in I$ \Comment{For each invalid group $I_e$}
  \State \ \ \ \ \ \ $\Phi \gets \Phi \cup \textsc{LearnPatterns}(I_e, V)$ \Comment{Learn patterns for group and add to filter set} 
  \State \Return $\Phi$
  \end{algorithmic}
\end{algorithm}

We now describe \name, our algorithm for learning prefix filters, and formally define 
what prefix filters actually are.
\name is based on the observation that, because the errors of an LLM are clustered 
around certain modes of failure, it is possible to identify and learn these clusters 
using a small set of unconstrained samples from the LLM: a process that we name as 
\emph{error-driven pattern learning}.

\Cref{alg:learning-alg-1} describes error-driven pattern learning: which, at a
high level, simply draws samples from $\aL$ (line 3),
divides them into valid and invalid samples (lines 4 and 5), then learns error patterns from the divided sets (line 7).
\Cref{alg:learning-alg-1} itself does not force that the patterns are prefix filters; it represents a general 
algorithm for learning the error patterns of LLMs.
The main idea in \Cref{alg:learning-alg-1} is that
many oracles support \emph{classifying} bad samples in depending on what the issue is; i.e., 
$\aO$ returns a tuple $(v, e)$ where $v$ indicates validity and $e$ indicates the error if 
$v=\Efalse$.
In lines 6 and 7, we group invalid samples into different sets 
$I_e$ depending on their error $e$; line 8 learns filters 
separately for each group.
The intuition is that information from the oracle can be exploited to further cluster invalid samples 
from the LLM $\aL$, which makes learning simpler filters easier; 
if an oracle does not have error support, \Cref{alg:learning-alg-1} can still proceed by placing all invalid samples in the 
same group.

\subsection{Prefix Filters}
\label{sec:prefix-filters}

\Cref{alg:learning-alg-1} defined a general algorithm for learning the error patterns of an LLM.
We now define and propose prefix filters 
as concrete representations of the error patterns learned by \textsc{LearnPatterns} in \Cref{alg:learning-alg-1}, 
that satisfies the need to 
\rone understand the error pattern of an LLM, and 
\rtwo efficiently constrain the output of an LLM at test time.

\begin{definition}[Prefix Filter]
\label{def:prefix-filter}
A \emph{prefix filter} is a function $\pfilter: \Sigma^* \rightarrow \set{\Etrue, \Efalse}$.
For a given string $s$, we say that a filter $\pfilter$ \emph{triggers} on $s$ if 
$\pfilter(s) = \Efalse$.

A filter $\pfilter$ is \emph{sound} with respect to an oracle $\aO$ iff
for all $s$ such that $\pfilter(s) = \Efalse$, given any suffix string $s'$, 
$\aO(s \cdot s') = \Efalse$---i.e.,  $\pfilter$
\emph{only} 
identifies prefixes that are \emph{guaranteed} to result in an invalid sample
for any suffix $s'$ according to the oracle $\aO$.

A filter $\pfilter$ is \emph{complete} with respect to an oracle $\aO$ iff 
for all $s$ such that for all $s'$, $\aO(s \cdot s') = \Efalse$, $\pfilter(s) = \Efalse$---i.e., 
$\pfilter$ recognizes \emph{all} samples $s$ that are guaranteed to result in an invalid completion.
\end{definition}

A prefix filter $\pfilter$ is simply a classifier on strings (prefixes of samples from an LLM).
Completeness of individual filters is \emph{not} a desideratum in our approach: 
our goal is to learn \emph{only} those filters that are relevant to the error pattern of an LLM, 
not those that recognize the entire space of invalid outputs.

On the other hand, soundness \emph{is} a desideratum for individual prefix filters: filters 
should not ban valid samples from being generated.
However, enforcing soundness precisely given an arbitrary oracle is often an undecidable task and results in very complex filters.
Instead, we exploit the fact that prefix filters are learned per-domain-and-LLM to define a notion of 
\emph{distributional soundness} instead.

\begin{definition}[Distributional Soundness]
\label{def:distributional-soundness}
Let $\aL$ be an LLM, $D$ a domain, $\aO$ an oracle for $D$, 
and $\Pr_D$ a distribution of prompts considered over $D$.
Let $\Pr_{\aL, D}$ define the distribution of \emph{outputs} $o$ obtained by
showing the LLM $\aL$ prompts $p$ sampled from $\Pr_D$:
$\Pr_{\aL, D}(o) = \sum_{p} \Pr_D(p) \cdot \Pr_{\aL}(o \mid p)$.

We say that a prefix filter $\pfilter$ is \emph{distributionally sound within $\epsilon$} 
with respect to an LLM $\aL$, a domain $D$, and an oracle $\aO$
iff for all samples $s$ such that $\pfilter(s) = \Efalse$ and suffixes $s'$ such that 
$\aO(s \cdot s') = \Etrue$, 
$\sum_{s \cdot s'} \Pr_{\aL, D}(s \cdot s') < \epsilon$: i.e.,
$\pfilter$ mis-triggers on a valid generation of $\aL$ 
with probability \emph{at most} $\epsilon$.
\end{definition}
Distributional soundness is a key design choice for prefix filters, allowing them to
\rone better approximate the boundary between valid and invalid outputs by 
sacrificing a small number of low-probability outputs, alleviating the misalignment problem, 
and
\rtwo be simpler and require less data to learn.

\begin{example}[Distributionally sound]
\label{ex:qwen-filter}
\name learns the following filter for TypeScript \& Qwen-1.5B:
\begin{lstlisting}[style=pyinline]
def detect_unnecessary_whitespace(seq_prefix: str) -> bool:
    s = seq_prefix.strip()
    return True if (not s) else False
\end{lstlisting}
This filter bans whitespace-led 
samples
and is unsound in general. 
However, this filter is distributionally sound on Qwen-1.5B, for which
whitespace-led samples are \emph{almost always} empty completions of a typed function body
which fail to compile;
the filter does not mis-reject valid samples.
\end{example}

In practice, 
we enforce distributional soundness by taking a small number of 
samples from the LLM $\aL$ to approximate the distribution $\Pr_{\aL, D}$.


\subsection{Learning and Constrained Generation with Prefix Filters}

\begin{algorithm}[t]
  \caption{\textsc{LearnFilters}: Querying an external LLM to learn prefix filters}
  \label{alg:learning-alg-2}
  \small
  \begin{algorithmic}[1]
  \State \textbf{Input:} Invalid samples $I_e$, valid samples $V$, retry limit $R$. \ \textbf{Output:} Set $\Phi$ of prefix filters for $I_e$
  \State $\Phi \gets \emptyset$
  \State ${\Phi_C} \gets \textsc{QueryLargeLLM}(I_e, V)$ \Comment{Query large LLM to get candidate filter set $\Phi_C$}
  \State \textbf{for each} ${\phi_C} \in {\Phi_C}$       \Comment{For each filter $\phi_C \in \Phi_C$}
  \State \textbf{for} $i$ \textbf{in range} $R$:
    \State \ \ \ \ \ \ $(v, E) \gets \textsc{Validate}(\phi_C, I_e, V)$     \Comment{Validate $\phi_C$}
    \State \ \ \ \ \ \ \textbf{if} $v$ \textbf{then break}                \Comment{If sound, break}
    \State \ \ \ \ \ \ \textbf{else} $\phi_C \gets \textsc{RequeryLargeLLM}(\phi_C, I_e, V, E)$ \Comment{Else requery LLM with error reason $E$}
  \State \textbf{if} $\textsc{Validate}(\phi_C, I_e, V)$ \textbf{then} $\Phi \gets \Phi \cup \set{\phi_C}$
    \Comment{If $\phi_C$ is sound, add $\phi_C$ to final set $\Phi$}
  \State \Return $\Phi$
  \end{algorithmic}
\end{algorithm}

\Cref{alg:learning-alg-2} illustrates how \name learns candidate filters.
\textsc{QueryLargeLLM} on line 4 first queries a large LLM on the given set of invalid and valid samples.
We then \textsc{Validate} each individual filter $\phi_C$ for 
\rone distributional soundness, and 
\rtwo whether the proposed filter captures invalid samples from $I_e$.
Validated filters are added to the filter set $\Phi$ (line 10); if validation fails, 
we requery the large LLM with additional information $E$ about why the filter $\phi_C$ failed validation 
(\textsc{RequeryLargeLLM}, line 9) up to a fixed number of times (line 6).
Filters that fail validation after the retry limit are discarded.

\paragraph{Constrained generation using prefix filters}
\begin{algorithm}[t]
\caption{CARS with Prefix Filters}
\label{alg:decoding-alg}
\begin{algorithmic}[1]
\State \textbf{Input:} LLM $\aL$, prompt $p$, prefix filters $\Phi$. \ \textbf{Output:} Sample $o$
\State $T \gets \emptyset$ \Comment{$T$ denotes token trie containing probabilities of token selection}
\State \textbf{while true}:
  \State \ \ \ \ \ \ $o \gets \epsilon, t \gets \epsilon$ \Comment{Set output $o$ and next token $t$ to $\epsilon$}
  \State \ \ \ \ \ \ \textbf{while} {$t \neq \text{EOF}$}
    \State \ \ \ \ \ \ \ \ \ \ \ \  $t \gets \textsc{SampleNextToken}(\aL, T, o); o \gets o \cdot t$ \Comment{Sample $t$ from $\aL$ using $T$}
    \State \ \ \ \ \ \ \ \ \ \ \ \  \textbf{if} $\exists \pfilter \in \mathcal{F}$ such that $\pfilter(o \cdot t)=\Efalse$:
      \Comment{If current prefix $o$ is caught by a prefix filter}
    \State \ \ \ \ \ \ \ \ \ \ \ \ \ \ \ \ \ \ $\textsc{UpdateTrie}(T, o)$; \textbf{break} 
      \Comment{Update trie $T$ to ban $o$ and restart sample}
  \State \ \ \ \ \ \ \Return $o$
\end{algorithmic}
\end{algorithm}
After obtaining a set of prefix filters $\Phi$ via \Cref{alg:learning-alg-2}, 
we use constrained adaptive rejection sampling (CARS)~\cite{cars} to constrain the output of $\aL$ 
when sampling.
\Cref{alg:decoding-alg} illustrates the decoding approach at a high level, where
\rone we reject a sample being generated as soon as the prefix matches a filter in $\Phi$, and 
\rtwo when rejecting the sample, we update the probability trie (i.e., the logits) of the currently 
explored node to zero, and rebalance the probability trie with respect to this mass reduction.
The latter rebalance allows subsequent sampling attempts to be more efficient;
we refer the reader to \cite{cars} for details.
For our purposes, what matters is that 
\rone prefix filters can be used directly in CARS to constrain the 
output of LLMs, and 
\rtwo prefix filters can be \emph{intersected} easily in CARS, by running every filter 
on the current prefix.

\section{Experimental Setup}
\label{sec:experiments}

We investigate how \name can learn error patterns 
as prefix filters in practice and how efficient the learned filters are for constrained 
generation across 5 local LLMs 
(Qwen2.5-1.5B, 7B, and 14B, Llama3.1-8B, and Gemma3-12B)
and 4 domains
(we refer the reader to \Cref{app:setup} for details): 

\textbf{MLIR: }The task is to generate compile-valid torch-MLIR functions.
Filters are learned from 50 samples, with the torch-MLIR compiler 
as the oracle.
We evaluate compile rate on a set of 200 samples from \Cref{alg:decoding-alg}, 
against unconstrained and grammar-constrained baselines.

\textbf{Molecules:} The task is to generate chemically valid molecules
(represented as SMILES strings~\cite{smiles})
with high QED (drug-likeliness) scores~\cite{qed}.
Filters are learned from 100 samples, using RDKit~\cite{rdkit} as the oracle for both validity and QED scores.
We evaluate validity rate and average QED scores on a set of 200 samples 
with unconstrained and grammar-constrained decoding as baselines.

\textbf{Dialogue summaries:} The task is to summarize 
dialogues from the HR-MultiWOZ dataset~\cite{hr-multiwoz} without leaking private data. 
Filters are learned from 500 samples (5 each for 100 dialogues), with a 
custom leak detector as the oracle.
We evaluate the rate of summaries with no leaks 
on a set of 500 samples (10 each on a disjoint set of 50 dialogues) 
against an unconstrained baseline.

\textbf{TypeScript:} The task is to solve TypeScript benchmarks from MultiPL-E~\cite{multiple}.
Filters are learned from 500 samples (5 each for 100 benchmarks), with 
the TypeScript compiler as the oracle.
We evaluate compile rate 
on a set of 500 samples 
(10 each for a disjoint set of 50 benchmarks) with unconstrained generation as the baseline.\footnote{
  We do not use grammar-constrained decoding as a baseline for TypeScript 
  because there is no publicly available grammar that fits our exact task
  (see \Cref{app:setup} for details).
}

We observe that 
for summary and TypeScript, the oracle captures domain validity rather than per-task correctness
(e.g., the TypeScript compiler catches compile errors, not whether a compiling solution solves the benchmark).
This is an intentional choice to learn filters for the entire domain, although in principle, 
we could learn filters specialized to a task as well.
  In \Cref{app:setup}, we also consider a fifth domain (SQL), which we do not include in the main text 
  due to overlap with other domains.

Prefix filters are learned by prompting
\textsc{QueryLargeLLM} in \Cref{alg:learning-alg-2} to synthesize prefix filters 
of type $\textsc{str} \rightarrow \textsc{bool}$
in Python.
For each domain-LLM pair, we performed \emph{three rounds} of filter learning: 
we first learn filters from unconstrained samples, 
then from the shifted distribution induced by the first round of filters, learn 
the second round of filters (same for the third round).
We perform multiple-round learning because filters can 
shift the error patterns of models and expose different error modes.
All rounds of learning used the same-sized learning set as described for each domain.

\section{Results}
\label{sec:results}
 \newcommand{\tallrow}{\rule{0pt}{4.4ex}}

  \newcommand{\covercell}[3]{%
    \cellcolor{cyan!#1!white}%
    \begin{tabular}{@{}c@{}}
      #2\%\\[-0.7ex]
      \makebox[2.8em]{\textcolor{black!35}{\dotfill}}\\[-0.8ex]
      {\scriptsize #3\%}
    \end{tabular}%
  }
\newcolumntype{Y}{>{\raggedright\arraybackslash}X}
 
 \begin{figure*}[t!]
  {\centering
  \small
  \setlength{\tabcolsep}{4pt}
  \renewcommand{\arraystretch}{1.08}

  \textbf{A. What filters does \name learn?}\\
  \vspace{1ex}
 {\scriptsize
    \begin{tabularx}{\textwidth}{@{}p{0.11\textwidth}p{0.12\textwidth}Yp{0.07\textwidth}@{}}
    \toprule
    \textbf{Domain} & \textbf{Model} & \textbf{What does the filter catch?} {\scriptsize\textcolor{black!55}{[learned round, error group]}} &
  Triggers\\
    \midrule
    MLIR &
    Qwen 7B &
    Undefined softmax-family op variants (e.g. \texttt{torch.aten.softmax})
    {\scriptsize\textcolor{black!55}{[r1, unknown op]}} &
    130 \\

    MLIR &
    Qwen 14B &
    Wrong arity for \texttt{torch.aten.addmm} {\scriptsize\textcolor{black!55}{[r1, operand count]}} &
    70 \\

    \cdashline{1-4}[0.4pt/1pt]

    Molecules &
    Llama 8B &
    Large overdecorated polyacyl scaffolds, a recurring low-QED molecular motif {\scriptsize\textcolor{black!55}{[r1,
  QED]}}&
    58 \\

    Molecules &
    Qwen 14B &
    Fused polycyclic aryl acetate/ester scaffolds with hydrophobic low-QED structure {\scriptsize\textcolor{black!
  55}{[r3, QED]}}&
    24 \\

    \cdashline{1-4}[0.4pt/1pt]

    Summary &
    Qwen 1.5B &
    Language that explicitly leaks names (e.g., ``their name is John'') or attributes an action to a name (e.g.,
  ``approved by John'')
      {\scriptsize\textcolor{black!55}{[r1, name leak]}} &
    179 \\

    Summary &
    Llama 8B &
    Patterns attributing a certain action to a person (e.g., ``approved by John'') {\scriptsize\textcolor{black!55}{[r1,
  name leak]}} &
    68 \\

    \cdashline{1-4}[0.4pt/1pt]

    TypeScript &
    Qwen 14B &
    Reassignment of variables qualified by \texttt{const} {\scriptsize\textcolor{black!55}{[r1, TS2588]}} &
    645 \\

    TypeScript &
    Qwen 1.5B &
    \texttt{import} declarations inside solutions {\scriptsize\textcolor{black!55}{[r1, TS1003]}}
    &
    543 \\
    \bottomrule
    \end{tabularx}
  }

  \vspace{1.0ex}

  \begin{minipage}[t]{0.43\textwidth}
  \textbf{B. How well do the filters catch errors?}\\[-0.4ex]

{\scriptsize
  \renewcommand{\arraystretch}{1.00}
  \begin{tabular*}{\linewidth}{@{\extracolsep{\fill}}llccccc@{}}
  \toprule
  \textbf{Domain} & & \textbf{Q1.5} & \textbf{Q7} & \textbf{Q14} & \textbf{L8} & \textbf{G12} \\
  \midrule
  MLIR & Full &
  \cellcolor{cyan!34!white}74.7 &
  \cellcolor{cyan!33!white}74.2 &
  \cellcolor{cyan!38!white}84.8 &
  \cellcolor{cyan!22!white}49.2 &
  \cellcolor{cyan!29!white}64.2 \\
  (catch \%) & Top-3 &
  \cellcolor{cyan!15!white}33.5 &
  \cellcolor{cyan!23!white}51.7 &
  \cellcolor{cyan!18!white}40.0 &
  \cellcolor{cyan!10!white}22.0 &
  \cellcolor{cyan!12!white}27.3 \\

  \cdashline{1-7}[0.4pt/1pt]

  Molec. & Full &
  \cellcolor{cyan!20!white}44.9 &
  \cellcolor{cyan!26!white}58.1 &
  \cellcolor{cyan!21!white}46.5 &
  \cellcolor{cyan!27!white}59.6 &
  \cellcolor{cyan!33!white}74.3 \\
  (catch \%) & Top-3 &
  \cellcolor{cyan!8!white}16.8 &
  \cellcolor{cyan!16!white}35.5 &
  \cellcolor{cyan!17!white}38.4 &
  \cellcolor{cyan!22!white}48.2 &
  \cellcolor{cyan!28!white}62.4 \\

  \cdashline{1-7}[0.4pt/1pt]

  Summ. & Full &
  \cellcolor{cyan!43!white}95.3 &
  \cellcolor{cyan!42!white}92.7 &
  \cellcolor{cyan!37!white}82.7 &
  \cellcolor{cyan!41!white}90.8 &
  \cellcolor{cyan!25!white}55.6 \\
  (catch \%) & Top-3 &
  \cellcolor{cyan!34!white}75.5 &
  \cellcolor{cyan!22!white}48.8 &
  \cellcolor{cyan!19!white}42.3 &
  \cellcolor{cyan!15!white}32.3 &
  \cellcolor{cyan!15!white}33.3 \\

  \cdashline{1-7}[0.4pt/1pt]

  TS & Full &
  \cellcolor{cyan!45!white}99.3 &
  \cellcolor{cyan!32!white}71.3 &
  \cellcolor{cyan!33!white}74.4 &
  \cellcolor{cyan!42!white}94.1 &
  \cellcolor{cyan!2!white}4.0 \\
  (catch \%) & Top-3 &
  \cellcolor{cyan!34!white}75.2 &
  \cellcolor{cyan!7!white}16.5 &
  \cellcolor{cyan!16!white}35.9 &
  \cellcolor{cyan!26!white}57.5 &
  \cellcolor{cyan!0!white}0.0 \\
  \bottomrule
  \end{tabular*}
  }
  \end{minipage}%
  \hfill
  \begin{minipage}[t]{0.53\textwidth}
  \textbf{C. How well do filters tranfer across models?}\\[-0.4ex]

 {\scriptsize
  \renewcommand{\arraystretch}{1.05}
  \begin{tabularx}{\linewidth}{@{}lcccY@{}}
  \toprule
  \textbf{Domain} & \textbf{Own} & \textbf{Cross} & \textbf{Ratio} & \textbf{Notes} \\
  \midrule
  MLIR &
  69.4\% &
  24.4\% &
  \cellcolor{orange!28!white}0.35 &
  Llama filters capture 45.5\% of Qwen 14B bad samples \\

  Molec. &
  56.7\% &
  13.9\% &
  \cellcolor{orange!35!white}0.24 &
  Gemma filters capture 43.2\% of Qwen 14B bad samples \\

  Summ. &
  83.4\% &
  68.1\% &
  \cellcolor{teal!22!white}0.82 &
  Name-leak filters transfer broadly across models \\

  TS &
  68.6\% &
  65.3\% &
  \cellcolor{teal!31!white}0.95 &
  Inflated cross-capture rate from distributional soundness \\
  \bottomrule
  \end{tabularx}
  }
%
%
%
  \end{minipage}

  \caption{
    Summary of filters learned by \name and how they interact with the error patterns of LLMs. 
    Panel A shows the most active filters per domain and how many times they 
    triggered during testing (\Cref{sec:learnability}).
    Panel B compares the percentage of unconstrained invalid samples  
    caught by all learned filters (top row), to the
    percentage of invalid samples that the top-3 active filters catch (bottom row), 
    showing how clustered the errors are (\Cref{sec:cluster}).
    Panel C investigates filter transfer, showing 
    the percentage of unconstrained samples a model's own filters catch (column Own),  compared with 
    the percentage of samples from \emph{other} models caught by the model's filters (column Cross)
    (\Cref{sec:transfer}).
    \Cref{app:filter-intervals} presents versions of Panel B and C with confidence intervals.
  }
  \label{fig:filter-table}
  }
  \end{figure*}

\Cref{fig:filter-table} presents a summarized analysis of the filters we learn, 
while \Cref{fig:efficacy-figure} presents results on constrained generation.
We center our evaluation around 4 main questions.
A detailed per-domain analysis of results can be found in \Cref{app:results}.
Comparisons made in this section are statistically supported 
at 95\% bootstrap confidence intervals following the method in~\cite{bootstrap}, 
unless otherwise noted.
In this section, we say that a filter has caught an error if it triggers 
on line 7 of \Cref{alg:decoding-alg}, meaning that the filter has prevented an invalid sample from being generated.


\subsection{Learnability: How well can \name distill error patterns as prefix filters?}
\label{sec:learnability}
Panel A in \Cref{fig:filter-table}
lists the most active filters 
for each domain: e.g., the Qwen-7B MLIR filter caught 130 invalid samples 
while generating the evaluation set of 200 MLIR samples via \Cref{alg:decoding-alg}.

Most active filters were learned at round 1,
although some active filters were learned during rounds 2 and 3 from the shifted error patterns.
For example, the Qwen-1.5B TypeScript filter was learned in round 2: the unconstrained distribution for TypeScript 
had 260 / 500 samples failing to compile with error code TS1005 (a syntax error).
The first round of filter learning shifted the distribution such that only 7 / 500 samples failed with code TS1005; 
the number of TS2304 errors, which the filter targets, increased from 23 to 51, 
allowing \textsc{QueryLargeLLM} to 
produce an effective filter.

Panel B illustrates that the learned filters are effective at catching invalid samples from the 
unconstrained distribution of the LLM:
for example, filters learned for 
MLIR \& Qwen-7B were able to detect 74.2\% of the unconstrained samples generated by the model, 
showing the filters are effective.
The metrics show that \name is indeed capable of learning prefix filters that correctly target error patterns, 
as shown by the high capture rates in Panel B; the one exception is 
TypeScript \& Gemma-12B, where prefix filters failed because Gemma almost always generated 
valid TypeScript (95.0\% compile rate) and there were too few invalid examples to learn filters from.

\textbf{Conclusion:} 
\name is capable of learning prefix filters that capture a wide portion of the per-domain-and-LLM error pattern.

\subsection{The cluster hypothesis: Do LLMs actually make a small set of clustered mistakes?}
\label{sec:cluster}
Panel B validates our cluster hypothesis: the top 3 filters for each domain-and-LLM 
catch a significant portion of errors.
For example, MLIR \& Qwen-7B learned a total of 20 filters, 
but the top 3 filters for Qwen-7B caught \emph{51.7\%} of the 
errors---in other words, the top three filters caught 70\% of the errors that the filters were able to catch.
An example filter for MLIR \& Qwen-7B is:
\begin{lstlisting}[style=pyinline]
def filter_unknown_torch_aten_softmax_family_op(text: str) -> bool:
    import re
    return re.search(
        r'(?m)^\s*%[\w.]+\s*=\s*torch\.aten\.(?:softmax|log_softmax|logsumexp)\s',
        text,
    ) is not None
\end{lstlisting}
This filter just rules out three concrete undefined MLIR operators 
(\texttt{torch.aten.softmax}, \texttt{log\_softmax}, or \texttt{logsumexp}), 
but triggers 
on 35.0\% of the errors from Qwen-7B.
Another example for Molecules \& Qwen-14B is:
\begin{lstlisting}[style=pyinline]
def filter_polycarbonyl_rich_scaffold(smiles_prefix: str) -> bool:
    return smiles_prefix.count("C(=O)") >= 3
\end{lstlisting}
This filter simply rejects molecules that already have 3 carbonyl (\texttt{C(=O)})
instances but triggers on 25.0\% of invalid attempts.
That these simple filters trigger on over 20\% of errors 
indicates that LLMs indeed have clustered error modes that can be detected and prevented 
with prefix filters.

\textbf{Conclusion:} LLMs do make clustered errors that can be caught with just a few filters.
There do remain errors outside of the clustered modes; prefix filters alone cannot capture the
entire set.

\subsection{Filter transferability: Can filters learned for one model be applied to another?}
\label{sec:transfer}
Panel C investigates how well filters learned from one model \emph{transfer} to another model: 
the `Own' column averages how well each model's own filters cover its unconstrained samples, 
while the `Other' column averages how well the filters cover samples from \emph{other} models.
MLIR and molecules display a very low ratio of transferability, illustrating that 
the error patterns of LLMs differ in these domains.
On the other hand, the summary domain shows a high level of transferability:
this is because across all models, name leaks are by far the most common mode of error.
For example,
\begin{lstlisting}[style=pyinline]
def filter_name_named_or_identified_as(seq_prefix: str) -> bool:
    return bool(re.search(
        r'\b(?:named|called|identified as|referred to as)\s+'
        r'[A-Z][a-z]+(?:\s+[A-Z][a-z]+){0,2}\b',
        seq_prefix,
    ))
\end{lstlisting}
This filter was learned for Qwen-1.5B, but applies broadly 
across all models as it identifies common name leak patterns: a lead-in (e.g., \texttt{named}), 
followed by the name, which first character is capitalized.

Results for TypeScript also suggest that the filters are broadly transferable, which is somewhat true,  
but TypeScript also displays special behavior in that 
distributionally sound filters for one model are not sound for others---thus being overly aggressive and inflating the 
trigger count.
For example, recall the \texttt{whitespace} filter for Qwen-1.5B discussed in \Cref{ex:qwen-filter}.
The filter is distributionally sound for Qwen-1.5B, but \emph{unsound} for Gemma, 
which often inserts newlines before valid solutions; 
this filter thus fires unconditionally on most Gemma samples and inflates the cross-validity ratio.

\textbf{Conclusion:} Filter transferability depends on the domain.
In some domains, LLMs are likely to make similar mistakes; in others, their error patterns are significantly different.

\subsection{Efficacy: How effective are the learned prefix filters for constrained generation?}
\begin{figure*}[t!]
{\centering
\small
\pgfplotsset{
  every axis/.append style={
    cycle list={{}},
  },
}

{\scriptsize
\begin{tabular}{ccccc}
\raisebox{0.35ex}{\tikz{\draw[SoftBlack, dotted, line width=0.7pt] (0,0) -- (0.35,0); \fill[SoftBlack, opacity=0.7] (0.175,0) circle (1.2pt);}} Qwen 1.5B &
\raisebox{0.35ex}{\tikz{\draw[SoftBlue, dotted, line width=0.7pt] (0,0) -- (0.35,0); \node[draw=SoftBlue, fill=SoftBlue, opacity=0.65, inner sep=1.1pt, regular polygon, regular polygon sides=4] at (0.175,0) {};}} Qwen 7B &
\raisebox{0.35ex}{\tikz{\draw[SoftGreen, dotted, line width=0.7pt] (0,0) -- (0.35,0); \node[regular polygon, regular polygon sides=3, fill=SoftGreen, opacity=0.7, inner sep=1.2pt] at (0.175,0) {};}} Qwen 14B &
\raisebox{0.35ex}{\tikz{\draw[BurntOrange, dotted, line width=0.7pt] (0,0) -- (0.35,0); \node[diamond, fill=BurntOrange, opacity=0.65, inner sep=1.2pt] at (0.175,0) {};}} Llama 8B &
\raisebox{0.35ex}{\tikz{\draw[RoyalPurple, dotted, line width=0.7pt] (0,0) -- (0.35,0); \node[regular polygon, regular polygon sides=5, fill=RoyalPurple, opacity=0.62, inner sep=1.1pt] at (0.175,0) {};}} Gemma 12B
\end{tabular}
}

\vspace{0.5ex}

\begin{minipage}[t]{0.235\textwidth}
\textbf{MLIR}\\[-1.2ex]

\begin{tikzpicture}
\begin{groupplot}[
  group style={
    group size=1 by 2,
    vertical sep=0.58cm,
    x descriptions at=edge bottom,
  },
  width=0.97\linewidth,
  height=0.145\textheight,
  xtick={1,2,3,4,5},
  xticklabels={U,G,R1,R2,R3},
  xticklabel style={font=\tiny},
  yticklabel style={font=\tiny},
  label style={font=\scriptsize},
  ylabel style={font=\scriptsize, yshift=-0.15em},
  grid=both,
  major grid style={black!6},
  minor grid style={black!3},
  tick align=outside,
]

\nextgroupplot[
  ylabel={Oracle validity (\%)},
  ymin=0,
  ymax=100,
  xticklabels={},
]
\addplot+[mark=*, mark size=1.5pt, line width=0.65pt, dotted, color=SoftBlack, opacity=0.70] coordinates {(1,9.0) (2,5.0) (3,13.5) (4,22.5) (5,31.5)};
\addplot+[mark=square*, mark size=1.5pt, line width=0.65pt, dotted, color=SoftBlue, opacity=0.65] coordinates {(1,11.0) (2,26.0) (3,32.0) (4,40.5) (5,43.0)};
\addplot+[mark=triangle*, mark size=1.7pt, line width=0.65pt, dotted, color=SoftGreen, opacity=0.70] coordinates {(1,27.5) (2,19.5) (3,46.5) (4,68.0) (5,71.0)};
\addplot+[mark=diamond*, mark size=1.7pt, line width=0.65pt, dotted, color=BurntOrange, opacity=0.65] coordinates {(1,34.0) (2,25.5) (3,38.0) (4,44.0) (5,49.0)};
\addplot+[mark=pentagon*, mark size=1.7pt, line width=0.65pt, dotted, color=RoyalPurple, opacity=0.62] coordinates {(1,6.5) (2,11.5) (3,10.5) (4,18.5) (5,22.5)};

\nextgroupplot[
  ylabel={Tok./sample},
  ymin=0,
  ymax=930,
]
\addplot+[mark=*, mark size=1.5pt, line width=0.65pt, dotted, color=SoftBlack, opacity=0.70] coordinates {(1,208.0) (2,355.3) (3,266.1) (4,305.6) (5,325.8)};
\addplot+[mark=square*, mark size=1.5pt, line width=0.65pt, dotted, color=SoftBlue, opacity=0.65] coordinates {(1,425.1) (2,842.8) (3,606.4) (4,657.9) (5,700.3)};
\addplot+[mark=triangle*, mark size=1.7pt, line width=0.65pt, dotted, color=SoftGreen, opacity=0.70] coordinates {(1,200.6) (2,202.8) (3,275.6) (4,360.0) (5,358.4)};
\addplot+[mark=diamond*, mark size=1.7pt, line width=0.65pt, dotted, color=BurntOrange, opacity=0.65] coordinates {(1,220.0) (2,233.8) (3,234.8) (4,274.7) (5,289.0)};
\addplot+[mark=pentagon*, mark size=1.7pt, line width=0.65pt, dotted, color=RoyalPurple, opacity=0.62] coordinates {(1,305.6) (2,330.7) (3,340.8) (4,483.2) (5,523.0)};

\end{groupplot}
\end{tikzpicture}
\end{minipage}%
\hfill
\begin{minipage}[t]{0.235\textwidth}
\textbf{Molecules}\\[-1.2ex]

\begin{tikzpicture}
\begin{groupplot}[
  group style={
    group size=1 by 2,
    vertical sep=0.58cm,
    x descriptions at=edge bottom,
  },
  width=0.97\linewidth,
  height=0.145\textheight,
  xtick={1,2,3,4,5},
  xticklabels={U,G,R1,R2,R3},
  xticklabel style={font=\tiny},
  yticklabel style={font=\tiny},
  label style={font=\scriptsize},
  ylabel style={font=\scriptsize, yshift=-0.15em},
  grid=both,
  major grid style={black!6},
  minor grid style={black!3},
  tick align=outside,
]

\nextgroupplot[
  ylabel={},
  ymin=0,
  ymax=100,
  xticklabels={},
]
\addplot[mark=*, mark size=1.5pt, line width=0.65pt, dotted, color=SoftBlack, opacity=0.70] coordinates {(1,39.5) (2,40.5) (3,45.5) (4,47.0) (5,48.0)};
\addplot[mark=square*, mark size=1.5pt, line width=0.65pt, dotted, color=SoftBlue, opacity=0.65] coordinates {(1,84.5) (2,85.5) (3,88.5) (4,89.0) (5,87.5)};
\addplot[mark=triangle*, mark size=1.7pt, line width=0.65pt, dotted, color=SoftGreen, opacity=0.70] coordinates {(1,83.5) (2,83.0) (3,84.5) (4,85.0) (5,85.0)};
\addplot[mark=diamond*, mark size=1.7pt, line width=0.65pt, dotted, color=BurntOrange, opacity=0.65] coordinates {(1,46.5) (2,53.0) (3,55.5) (4,61.0) (5,62.0)};
\addplot[mark=pentagon*, mark size=1.7pt, line width=0.65pt, dotted, color=RoyalPurple, opacity=0.62] coordinates {(1,93.5) (2,93.0) (3,91.5) (4,93.0) (5,93.0)};

\nextgroupplot[
  ylabel={},
  ymin=0,
  ymax=65,
]
\addplot[mark=*, mark size=1.5pt, line width=0.65pt, dotted, color=SoftBlack, opacity=0.70] coordinates {(1,28.2) (2,28.0) (3,31.4) (4,35.0) (5,36.2)};
\addplot[mark=square*, mark size=1.5pt, line width=0.65pt, dotted, color=SoftBlue, opacity=0.65] coordinates {(1,28.6) (2,29.0) (3,30.7) (4,34.9) (5,40.3)};
\addplot[mark=triangle*, mark size=1.7pt, line width=0.65pt, dotted, color=SoftGreen, opacity=0.70] coordinates {(1,26.6) (2,26.7) (3,28.4) (4,29.0) (5,33.5)};
\addplot[mark=diamond*, mark size=1.7pt, line width=0.65pt, dotted, color=BurntOrange, opacity=0.65] coordinates {(1,47.1) (2,49.6) (3,55.9) (4,57.8) (5,58.4)};
\addplot[mark=pentagon*, mark size=1.7pt, line width=0.65pt, dotted, color=RoyalPurple, opacity=0.62] coordinates {(1,22.2) (2,22.3) (3,26.0) (4,28.4) (5,30.7)};

\end{groupplot}
\end{tikzpicture}
\end{minipage}%
\hfill
\begin{minipage}[t]{0.235\textwidth}
\textbf{Summary}\\[-1.2ex]

\begin{tikzpicture}
\begin{groupplot}[
  group style={
    group size=1 by 2,
    vertical sep=0.58cm,
    x descriptions at=edge bottom,
  },
  width=0.97\linewidth,
  height=0.145\textheight,
  xtick={1,2,3,4},
  xticklabels={U,R1,R2,R3},
  xticklabel style={font=\tiny},
  yticklabel style={font=\tiny},
  label style={font=\scriptsize},
  ylabel style={font=\scriptsize, yshift=-0.15em},
  grid=both,
  major grid style={black!6},
  minor grid style={black!3},
  tick align=outside,
]

\nextgroupplot[
  ylabel={},
  ymin=0,
  ymax=100,
  xticklabels={},
]
\addplot[mark=*, mark size=1.5pt, line width=0.65pt, dotted, color=SoftBlack, opacity=0.70] coordinates {(1,78.8) (2,95.6) (3,96.0) (4,97.2)};
\addplot[mark=square*, mark size=1.5pt, line width=0.65pt, dotted, color=SoftBlue, opacity=0.65] coordinates {(1,75.4) (2,89.6) (3,92.6) (4,94.2)};
\addplot[mark=triangle*, mark size=1.7pt, line width=0.65pt, dotted, color=SoftGreen, opacity=0.70] coordinates {(1,89.6) (2,96.0) (3,96.6) (4,96.6)};
\addplot[mark=diamond*, mark size=1.7pt, line width=0.65pt, dotted, color=BurntOrange, opacity=0.65] coordinates {(1,87.0) (2,96.2) (3,98.0) (4,98.0)};
\addplot[mark=pentagon*, mark size=1.7pt, line width=0.65pt, dotted, color=RoyalPurple, opacity=0.62] coordinates {(1,98.2) (2,98.4) (3,98.8) (4,98.8)};

\nextgroupplot[
  ylabel={},
  ymin=0,
  ymax=300,
]
\addplot[mark=*, mark size=1.5pt, line width=0.65pt, dotted, color=SoftBlack, opacity=0.70] coordinates {(1,159.2) (2,197.4) (3,202.2) (4,203.6)};
\addplot[mark=square*, mark size=1.5pt, line width=0.65pt, dotted, color=SoftBlue, opacity=0.65] coordinates {(1,143.1) (2,228.0) (3,244.4) (4,248.7)};
\addplot[mark=triangle*, mark size=1.7pt, line width=0.65pt, dotted, color=SoftGreen, opacity=0.70] coordinates {(1,213.4) (2,241.5) (3,242.7) (4,244.0)};
\addplot[mark=diamond*, mark size=1.7pt, line width=0.65pt, dotted, color=BurntOrange, opacity=0.65] coordinates {(1,238.9) (2,260.4) (3,264.6) (4,269.3)};
\addplot[mark=pentagon*, mark size=1.7pt, line width=0.65pt, dotted, color=RoyalPurple, opacity=0.62] coordinates {(1,72.9) (2,74.5) (3,74.6) (4,75.6)};

\end{groupplot}
\end{tikzpicture}
\end{minipage}%
\hfill
\begin{minipage}[t]{0.235\textwidth}
\textbf{TypeScript}\\[-1.2ex]

\begin{tikzpicture}
\begin{groupplot}[
  group style={
    group size=1 by 2,
    vertical sep=0.58cm,
    x descriptions at=edge bottom,
  },
  width=0.97\linewidth,
  height=0.145\textheight,
  xtick={1,2,3,4},
  xticklabels={U,R1,R2,R3},
  xticklabel style={font=\tiny},
  yticklabel style={font=\tiny},
  label style={font=\scriptsize},
  ylabel style={font=\scriptsize, yshift=-0.15em},
  grid=both,
  major grid style={black!6},
  minor grid style={black!3},
  tick align=outside,
]

\nextgroupplot[
  ylabel={},
  ymin=0,
  ymax=100,
  xticklabels={},
]
\addplot[mark=*, mark size=1.5pt, line width=0.65pt, dotted, color=SoftBlack, opacity=0.70] coordinates {(1,16.0) (2,51.6) (3,61.4) (4,70.0)};
\addplot[mark=square*, mark size=1.5pt, line width=0.65pt, dotted, color=SoftBlue, opacity=0.65] coordinates {(1,77.0) (2,89.8) (3,88.2) (4,87.4)};
\addplot[mark=triangle*, mark size=1.7pt, line width=0.65pt, dotted, color=SoftGreen, opacity=0.70] coordinates {(1,92.2) (2,92.8) (3,93.8) (4,93.0)};
\addplot[mark=diamond*, mark size=1.7pt, line width=0.65pt, dotted, color=BurntOrange, opacity=0.65] coordinates {(1,45.4) (2,65.4) (3,65.4) (4,66.6)};
\addplot[mark=pentagon*, mark size=1.7pt, line width=0.65pt, dotted, color=RoyalPurple, opacity=0.62] coordinates {(1,95.0) (2,95.0) (3,95.0) (4,95.2)};

\nextgroupplot[
  ylabel={},
  ymin=0,
  ymax=900,
]
\addplot[mark=*, mark size=1.5pt, line width=0.65pt, dotted, color=SoftBlack, opacity=0.70] coordinates {(1,46.1) (2,226.2) (3,377.8) (4,467.0)};
\addplot[mark=square*, mark size=1.5pt, line width=0.65pt, dotted, color=SoftBlue, opacity=0.65] coordinates {(1,58.8) (2,90.5) (3,97.3) (4,107.6)};
\addplot[mark=triangle*, mark size=1.7pt, line width=0.65pt, dotted, color=SoftGreen, opacity=0.70] coordinates {(1,133.7) (2,267.0) (3,272.2) (4,270.6)};
\addplot[mark=diamond*, mark size=1.7pt, line width=0.65pt, dotted, color=BurntOrange, opacity=0.65] coordinates {(1,167.7) (2,468.5) (3,805.7) (4,838.8)};
\addplot[mark=pentagon*, mark size=1.7pt, line width=0.65pt, dotted, color=RoyalPurple, opacity=0.62] coordinates {(1,95.3) (2,95.4) (3,95.4) (4,95.3)};

\end{groupplot}
\end{tikzpicture}
\end{minipage}

\caption{
Validity and token costs per sample across domains and LLMs.
The first row plots validity rates according to the oracle for each domain.
X-axis ticks are ordered (U)nconstrained, (G)rammar-constrained (for MLIR and Molecules), and 
(R)ounds 1, 2, and 3 of filters.
An upward right trend indicates that validity (top row) / tokens per sample (bottom row) is increasing as filters are added.
}
\label{fig:efficacy-figure}
}
\end{figure*}
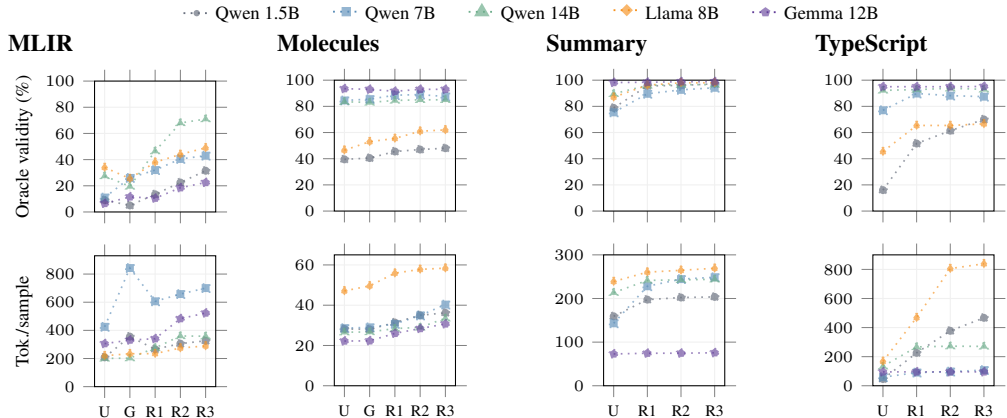

%
%
%
%

Next, we investigate how effective the learned filters are for constraining the
generation of LLMs towards valid samples (RQ4).
\Cref{fig:efficacy-figure} summarizes the results for RQ4: 
the top row plots the ratio of samples that the oracle accepts as valid 
(compilable functions for MLIR and TypeScript, chemically valid molecules for Molecules, and 
summaries without leaks for Summary)
while the bottom row plots the number of tokens required to produce each sample.

\paragraph{Validity gains.}
In most cases, oracle validity shows a clear increase in filter-constrained generation compared to unconstrained, 
or even grammar-constrained decoding: 
e.g., TypeScript validity jumps from 16.0\% to 70.0\% after all three rounds of filter learning.
The graphs also illustrate that for most cases, each round of filter learning helps increase the 
validity rate (e.g., for MLIR). 
MLIR is a domain where the gains from prefix filters are unique: 
grammar-constrained decoding suffers for MLIR because grammars cannot prevent prevalent 
error modes such as using undefined operators.
Prefix filters prevent such error modes and achieve a much higher compile rate.

The exception to validity rate increase is when the model becomes competent enough that 
there are too few invalid examples to learn invalid filters.\footnote{
  Validity increase for Molecules \& Qwen-1.5B is statistically insignificant at the 95\% interval, 
  but becomes significant at 90\%.
  All other experiments saw an increase when the unconstrained model had $<80\%$ validity.
}
For example, Gemma-12B shows a 95.0\% validity rate unconstrained for TypeScript, which resulted in 
only 25 invalid samples for learning filters; the few filters suggested by 
\textsc{QueryLargeLLM} failed to improve the distribution significantly.
Alternatively, there is one case where the filters merely shift the pattern towards a different 
set of errors without increasing the validity rate, as was the case for 
TypeScript \& Llama-8B on between different filter rounds 
(the filters still do significantly increase validity compared to unconstrained).

We note that the prefix filters are not increasing validity rates by sacrificing other quality metrics.
For example, in the Summary domain, 
the average number of non-protected data fields 
each summary contained remained stable, indicating that the filters precisely targeted 
only prefixes that led to actual leaks. 
For TypeScript, the number of benchmarks at pass@10 increased significantly for 
Qwen-1.5B ($16 \rightarrow 29$ solved, out of 50), while showing a minor increase or staying stable for 
other models; this shows that the prefix filters are not simply collapsing the samples towards functions 
that compile trivially.
Detailed per-domain results including secondary quality metrics can be found in \Cref{app:results}.

\textbf{Conclusion:} For cases where there were enough invalid samples to learn effective prefix filters, 
prefix filters are highly effective at constraining the output of the LLM towards valid outputs.

\textbf{Overhead.}
Filter-constrained generation shows a strict increase in required tokens compared to unconstrained 
generation, which is expected due to the nature of rejection sampling (\Cref{alg:decoding-alg}).
More importantly, the increase in tokens per sample is relatively modest for most domains, compared to the gains:
MLIR had an average increase of 60.5\% in tokens per sample, 
molecules had an average increase of 31.3\%, and 
the summaries domain had an average increase of 26.5\%.

The one domain where tokens per sample suffered significantly was TypeScript, 
where Qwen-1.5B and Llama 8B incurred an overhead of 914.1\% / 400.2\% respectively.
For Qwen-1.5B, the high overhead is because the model is weak: 
the prefix filters help prevent invalid samples, but \Cref{alg:decoding-alg} 
must still pay a cost for each mistake the model makes.
For Llama-8B, the issue is that the model often makes mistakes at the \emph{tail} of samples, e.g., 
by adding non-code explanations after the solution.
Although Llama-8B makes fewer mistakes than Qwen-1.5B, this specific pattern of errors is very expensive 
for \Cref{alg:decoding-alg}, and overhead becomes high.

\textbf{Conclusion:} Tokens per sample overhead remains modest in general, but can become higher in a few cases
that prefix filters are not well-suited for.

\section{Limitations and Discussion}
\label{sec:discussion}

\paragraph{Limitations.}
One limitation of prefix filters and \name
is that they are ineffective on LLMs that are already competent on a domain, and 
generate few invalid examples from which we can
learn effective prefix filters.
This problem can be somewhat alleviated by 
generating a larger learning set for filter learning, 
or perhaps learning specialized filters tailored for a specific task within the domain, 
but both approaches also increase the overall cost of \name.
There are also some cases where prefix filters do not translate into an immediate gain 
in validity (e.g., TypeScript \& Llama-8B), and merely shift the error pattern towards
different errors.
It is unclear whether continued rounds of filter learning will eventually improve 
validty rates, or whether the errors will just become dispersed, in these cases.

Another limitation is that prefix filters may be unlearnable if the error 
is not soundly prefix-expressible: e.g., if 
\texttt{list.print} is not a valid operator but \texttt{list.printAll} is, then \name may be unable to 
learn prefix filters that prevent \texttt{list.print} from occurring.
An example is illustrated in \Cref{app:results}.

\paragraph{Related work.}
There are several other bodies of related work that attempt to solve the constrained generation problem 
outside of constrained decoding and fine-tuning.
Verbalized RL~\cite{reflexion} 
translates outputs from oracles into text that is prompted to the LLM on subsequent 
queries, attempting to prevent the LLM from making similar mistakes via prompting.
Some variants of grammar-constrained decoding~\cite{gcd} have elements that attempt to learn the grammar 
automatically, similar to how we learn prefix filters.
In contrast, our main focus is on analyzing the error patterns of LLMs \emph{themselves}; 
the application towards constrained generation is an example of how this information can be exploited.


\paragraph{Future work.}
 \name and prefix filters open several directions for future work:
 for example, generating error-focused datasets for fine-tuning, 
 potentially removing recurring failure modes with less data and compute. 
 Another possible extension is towards agentic systems, which often 
 repeat mistakes such as malformed tool calls. 
 Instead of relying on natural-language memory files that grow,
  consume context, and degrade over long horizons, 
  agents could learn prefix filters from repeated failures 
  and use them to prevent similar errors without adding prompt context.



\section{Conclusion}
We proposed \emph{prefix filters}, symbolic functions that 
capture per-domain-and-LLM error patterns, 
and \name, a system for learning prefix filters.
We showed that prefix filters can be learned and used to 
constrain LLM efficiently in practice; prefix filters 
provided insight into the error patterns of different LLMs, 
showing that different LLMs make different mistakes in many cases.
We anticipate that prefix filters have many future applications such as 
fine-tuning or making agentic behavior reliable.

\bibliography{reference}
\bibliographystyle{plainnat}


\newpage
\appendix
\section{Model Checkpoints}
We used the following models in our evaluation:\\

 \begin{tabular}{lll}
  \toprule
  Model & URL & Commit \\
  \midrule
  Qwen 1.5B &
  \url{https://huggingface.co/Qwen/Qwen2.5-1.5B-Instruct} &
  \texttt{989aa7} \\
  Qwen 7B &
  \url{https://huggingface.co/Qwen/Qwen2.5-7B-Instruct} &
  \texttt{a09a35} \\
  Qwen 14B &
  \url{https://huggingface.co/Qwen/Qwen2.5-14B-Instruct} &
  \texttt{cf98f3} \\
  Llama 8B &
  \url{https://huggingface.co/meta-llama/Llama-3.1-8B-Instruct} &
  \texttt{0e9e39} \\
  Gemma 12B &
  \url{https://huggingface.co/google/gemma-3-12b-it} &
  \texttt{96b6f1} \\
  \bottomrule
  \end{tabular}

For the large LLM in \textsc{QueryLargeLLM}, we queried 
OpenAI gpt-5.4-mini, accessed April 21--30, 2026.

\section{Hardware and Software}
\label{app:hardware}
  Experiments were conducted on a Ubuntu 20.04 LTS server equipped with a AMD EPYC 7282 CPU
  (32 physical cores total at up to 2.8GHz), 512GB of RAM, and NVIDIA LS40 GPUs.
  Each experiment was run on a single GPU. 
  We note that \name or constrained sampling using prefix filters does not require additional hardware outside 
  of a standard CPU; the GPU was used only for local model inference.
  Our implementation is based on Python 3.12.12, PyTorch 2.10.0 with CUDA
  12.8, HuggingFace Transformers 5.1.0, RDKit 2025.9.5, and torch-mlir 20260314.751.

\section{Details on the experimental setup}
\label{app:setup}
We provide more details on each domain of our experimental setup.
We enforce a 1\% error rate for distributional soundness, both on the model's own distribution, 
and also on a external dataset of validated solutions for each task.
We find that augmenting the model's own distribution with an external solution set helps 
block filters that are distributionally sound on the model's distribution, but would trigger on
candidate solutions that arise from shifted distribution obtained post-filtering.

\paragraph{MLIR}
For MLIR, the local LLM is tasked with generating a valid (compilable) MLIR function without 
additional constraints.
Our oracle for this task is the MLIR compiler \texttt{torch-mlir-opt}, which returns $\Etrue$ if the 
sample successfully compiles and $\Efalse$ (along with an error message) if the sample fails 
to compile.
We sample 50 MLIR programs for the learning filters and 200 for evaluating constrained generation 
on the learned filters.
Learned filters are validated on valid learning samples generated by the local LLM in addition to 
a fixed set of valid MLIR programs.
Our baselines for this domain are unconstrained generation, as well as grammar-constrained decoding 
using a Lark grammar we adapted from one previously used in other projects on 
constrained decoding~\cite{mlirgrm,germinator,cars}.

\emph{Error Classes.}
\texttt{torch-mlir-opt}, like many other compilers, emits an error message detailing why the 
program cannot compile.
If a learning sample is invalid, we take the \emph{first} error message emitted by 
\texttt{torch-mlir-opt} and use it to classify sample into one of 8 error classes 
(use of undefined custom operators, operator arity errors, type errors, SSA errors, syntax errors, 
invalid type annotations, invalid literals, and other miscellaneous errors).

\paragraph{Molecules}
For molecules, the local LLM is tasked with generating a chemically valid molecule in the SMILES 
string format without additional constraints.
We sample 100 molecules for the learning phase and sample 200 molecules after learning prefix filters as the testing 
phase.

We use \emph{two} oracles for the molecules task: 
\rone RDKit~\cite{rdkit}, which returns a binary $\Etrue$ / $\Efalse$ indicating whether 
the molecule is chemically valid, and 
\rtwo QED score~\cite{qed}, which is a real value between $0$ and $1$ indicating how good the sample is 
as a drug.
QED score is converted as a binary oracle for filter learning by analyzing the distribution of QED scores 
generated by the model, and setting those with a higher percentile as good samples, while 
setting those with a lower percentile as bad samples.
Learned filters are validated against valid (chemically valid / high QED-scoring) learning samples 
and an external set of 1000 molecules from the MOSES SMILES dataset~\cite{moses}.
Our baselines for this domain are unconstrained generation, as well as grammar-constrained decoding 
using a SMILES grammar adapted from GenLM~\cite{smilesgrm,genLM}.

We set both QED score and validity as filter targets to demonstrate that 
\name can learn prefix filters that optimize over multiple targets at once, not just a single target.
We note that the authors are unfamiliar with molecular chemistry: the molecules domain 
also serves as an example of how \name can be used without any domain expertise to enhance the performance of 
LLMs.

\emph{Error Classes.}
Similar to a program compiler, RDKit provides justification on why a given sample is 
chemically invalid 
(e.g., a chemical ring is never closed); we use this information to classify chemically invalid 
samples into 8 categories.
For QED score, all `bad' samples (low-QED scoring molecules) share a single error group.

\paragraph{Summaries}
In the summaries domain, the local LLM is given as input a dialogue between an HR department and a worker 
and asked to summarize the dialogue without leaking any private information.
The HR dialogues are took from the HR-MultiWOZ dataset~\cite{hr-multiwoz}, 
consisting of 550 HR dialogues, each of which are 
annotated with information fields that the dialogue contains 
(e.g., name, email, phone).
We define private information for this task as the fields \texttt{name}, \texttt{phone}, \texttt{id}, \texttt{email}, 
and \texttt{contact\_bundle} (which contains miscellaneous contact info); 
other fields exist in the dataset as well (e.g., \texttt{time\_off}, \texttt{training\_request}, etc.).
The goal of the local LLM is to generate a summary of the dialogue that does \emph{not} contain any of the 
5 protected fields but contains as many of the other fields as possible.
The oracle takes the values of the protected fields within the dataset 
and returns $\Efalse$ if any of the concrete values are present within the generated summary.

We take 100 dialogues from the dataset 
and generate 5 summary samples per dialogue for learning, then take a disjoint set of 
50 dialogues and generate 10 samples each for testing constrained generation 
(thus keeping the learning and testing sets separate).
Learned filters are validated against the non-leaking summaries generated by the local model as well 
as a set of 388 safe reference summaries generated by GPT5.5 and validated externally.
We use only unconstrained generation as a baseline, as it is hard to develop a formal grammar for 
natural language summaries; the summaries domain illustrates that prefix filters can be learned 
and applied to constrained generation even in such fuzzy domains.

\emph{Error Classes.}
Invalid samples in the HR domain (those that leak sensitive information) are classified according to 
which field of sensitive information they leak for a total of 5 groups.
Note that a single sample can leak multiple fields at once, in which case the offending sample will 
be placed in \emph{all} relevant buckets, not just one.

\paragraph{TypeScript}
For TypeScript, we take the MultiPL-E benchmark~\cite{multiple}, which contains 390 benchmark entries, 
each with the goal of synthesizing a TypeScript function that meets a natural-language specification.
The task of the local LLM is to solve these benchmarks.
Our oracle for this domain is the TypeScript compiler, which returns $\Etrue$ if the generated sample 
successfully compiles and $\Efalse$ otherwise.
We do \emph{not} take whether the local LLM succeeded in generating a \emph{correct} solution for a given problem 
as our validity signal---doing so would learn prefix filters that are specialized for solving a particular 
TypeScript problem, instead of generating valid TypeScript in general.

We take 100 problems from the MultiPL-E benchmark and generate 5 samples each for learning prefix filters, then 
take a disjoint set of 50 problems and generate 10 samples each for testing constrained generation.
Learned filters are validated against the compiling learning samples generated by the local LLM as well 
as an external set of 390 solutions for the MultiPL-E dataset.
We take only unconstrained generation as the baseline for TypeScript.

We do not compare against grammar-constrained decoding, despite the fact that TypeScript has been
studied previously in grammar-constrained settings, for two main reasons.
First, as far as the authors are aware, there is no official TypeScript grammar; 
the official specification of TypeScript is given via the implementation of the TypeScript compiler instead.
Second, while there do exist third-party grammars (e.g.,~\cite{treesitter}) for TypeScript, 
these grammars are not directly aligned with our benchmark tasks of completing 
a TypeScript function whose signature has already been defined; our implementation 
also depends specifically on Lark~\cite{lark}, and also includes prompting with triple-quote-fencing (\texttt{```}) 
to increase the performance of the local LLMs.
While it would be possible in theory to adapt existing implementations towards our particular scenario, 
doing so would require nontrivial effort.
\name and prefix filters propose a path towards constrained generation \emph{without} needing such effort, 
which is why we do not compare against grammar-constrained decoding for TypeScript.

\emph{Error Classes.}
Invalid TypeScript samples are categorized by the first error code that the 
TypeScript compiler emits~\cite{tserrors}.
There are many TypeScript error codes which may lead to a very large set of error classes, but in practice 
we find that invalid samples from the LLM exercise just a few 
(e.g., TS1005: token expected or TS2339: property does not exist on type), keeping the set of error classes 
small and relatively concentrated.

\paragraph{SQL}
We introduce SQL queries as an additional domain, which is not presented in the main text due to 
overlaps with the other domains.
For the SQL domain, the goal is to synthesize valid SQL queries that solve a certain natural-language 
SQL task, such as retrieving a certain value.
Our dataset is the \texttt{formula\_1} database from the BIRD-Mini Dev PostgreSQL benchmark~\cite{bird}, 
a 13-table database with 94 columns, 19 foreign-key links, and roughly 493000 rows, paired with
66 query tasks over the \texttt{formula\_1} database also from BIRD.
Our validity oracle is the PostgreSQL execution engine.

We use 26 tasks for learning filters and generate 5 samples per learning task for a total of 130 learning samples, 
and the remaining 40 tasks with 10 samples per task for a total of 400 testing samples.
Like TypeScript, we learn filters targeting the validity of a query itself, not whether the query solves a specific task, 
in order to learn filters that are broadly applicable to the entire domain.
Candidate filters are validated against the model's own learning set plus an external solution set of 400 valid 
SQL queries.
Our baselines for the SQL domain are unconstrained generation and grammar-constrained decoding.

\emph{Error Classes.}
The PostgreSQL engine also returns a variety of errors for invalid queries like the TypeScript compiler.
For the SQL domain, we group the errors into seven groups: 
parse errors, scope errors, aggregation / grouping errors, function signature errors, type errors, 
queries that attempt to access a missing relation or column, and other miscellaneous errors.

\section{Prompts}
\label{app:prompts}

Prompts to query the small / large LLMs are included as supplementary material with 
our submission.

\section{Extra per-domain experimental results}
\label{app:results}
We provide a detailed per-domain analysis of our experimental results in \Cref{sec:results}.

\subsection{MLIR}
  \begin{figure*}[t!]
    {\centering
    \small
    \setlength{\tabcolsep}{4pt}
    \renewcommand{\arraystretch}{1.08}

    \begin{minipage}[t]{0.52\textwidth}
    \textbf{A. Most-triggered MLIR prefix filters per-model}\\[-0.5ex]

  \begin{tabular}{@{}p{0.13\linewidth}p{0.82\linewidth}}
  \toprule
  Model & Top triggered filters {\scriptsize\textcolor{black!55}{[learned round]}} \\
  \midrule
  Qwen 1.5B &
    \triggerbar{Malformed \texttt{func.func} headers / bad sigils}{r0}{5.8}{63}{18.2}
  \\[-1ex]
  &
    \triggerbar{Invented custom-looking \texttt{torch.*custom*} ops}{r0}{4.8}{52}{15.0}
  \\[-0.8ex]
  \multicolumn{2}{@{}l}{\textcolor{black!40}{\makebox[\linewidth][l]{\dotfill}}}\\[-0.4ex]

  Qwen 7B &
    \triggerbar{Unsupported softmax-family op forms}{r0}{10.6}{130}{33.2}
  \\[-1ex]
  &
    \triggerbar{Unsupported miscellaneous ATen op names}{r0}{4.6}{57}{14.5}
  \\[-0.8ex]
  \multicolumn{2}{@{}l}{\textcolor{black!40}{\makebox[\linewidth][l]{\dotfill}}}\\[-0.4ex]

  Qwen 14B &
    \triggerbar{Wrong arity for \texttt{torch.aten.addmm}}{r0}{7.4}{70}{23.2}
  \\[-1ex]
  &
    \triggerbar{Wrong arity for other fixed-arity ATen ops}{r1}{3.5}{33}{10.9}
  \\[-0.8ex]
  \multicolumn{2}{@{}l}{\textcolor{black!40}{\makebox[\linewidth][l]{\dotfill}}}\\[-0.4ex]

  Llama 8B &
    \triggerbar{Additional unsupported ATen op names}{r1}{4.3}{14}{13.3}
  \\[-1ex]
  &
    \triggerbar{Invented or unsupported Torch op names}{r0}{3.6}{12}{11.4}
  \\[-0.8ex]
  \multicolumn{2}{@{}l}{\textcolor{black!40}{\makebox[\linewidth][l]{\dotfill}}}\\[-0.4ex]

  Gemma 12B &
    \triggerbar{Unsupported exact custom op names}{r0}{4.6}{48}{14.4}
  \\[-1ex]
  &
    \triggerbar{Too many operands for known fixed-arity ops}{r0}{4.5}{47}{14.1}
  \\
  \bottomrule
  \end{tabular}
    \end{minipage}%
    \hfill
    \begin{minipage}[t]{0.44\textwidth}
  \textbf{B. MLIR performance by method}\\[-1.3ex]

  {\scriptsize
  \setlength{\tabcolsep}{2.4pt}
  \begin{tabular}{@{}ccc@{}}
  \raisebox{0.35ex}{\tikz{\draw[SoftBlack, dotted, line width=0.7pt, opacity=0.70] (0,0) -- (0.35,0); \fill[SoftBlack, opacity=0.70] (0.175,0) circle (1.2pt);}} Qwen 1.5B &
  \raisebox{0.35ex}{\tikz{\draw[SoftBlue, dotted, line width=0.7pt, opacity=0.65] (0,0) -- (0.35,0); \node[draw=SoftBlue, fill=SoftBlue, opacity=0.65, inner sep=1.1pt, regular
  polygon, regular polygon sides=4] at (0.175,0) {};}} Qwen 7B &
  \raisebox{0.35ex}{\tikz{\draw[SoftGreen, dotted, line width=0.7pt, opacity=0.70] (0,0) -- (0.35,0); \node[regular polygon, regular polygon sides=3, fill=SoftGreen, opacity=0.70,
  inner sep=1.2pt] at (0.175,0) {};}} Qwen 14B \\[-0.2ex]
  \raisebox{0.35ex}{\tikz{\draw[BurntOrange, dotted, line width=0.7pt, opacity=0.65] (0,0) -- (0.35,0); \node[diamond, fill=BurntOrange, opacity=0.65, inner sep=1.2pt] at (0.175,0)
  {};}} Llama 8B &
  \raisebox{0.35ex}{\tikz{\draw[RoyalPurple, dotted, line width=0.7pt, opacity=0.62] (0,0) -- (0.35,0); \node[regular polygon, regular polygon sides=5, fill=RoyalPurple,
  opacity=0.62, inner sep=1.1pt] at (0.175,0) {};}} Gemma 12B &
  \end{tabular}
  }\\[-0.8ex]

  \begin{tikzpicture}
  \begin{groupplot}[
    group style={group size=1 by 2, vertical sep=0.6cm, x descriptions at=edge bottom},
    width=0.96\linewidth,
    height=0.42\linewidth,
    xtick={1,2,3,4,5},
    xticklabels={U,Gr,R1,R2,R3},
    xticklabel style={font=\tiny},
    yticklabel style={font=\tiny},
    label style={font=\scriptsize},
    ylabel style={font=\scriptsize, yshift=-0.15em},
    grid=both,
    major grid style={black!6},
    minor grid style={black!3},
    tick align=outside,
  ]

  \nextgroupplot[ylabel={Compile (\%)}, ymin=0, ymax=76, xticklabels={}]
  \addplot[mark=*, mark size=1.5pt, line width=0.65pt, dotted, color=SoftBlack, opacity=0.70] coordinates {(1,9.0) (2,5.0) (3,13.5) (4,22.5) (5,31.5)};
  \addplot[mark=square*, mark size=1.5pt, line width=0.65pt, dotted, color=SoftBlue, opacity=0.65] coordinates {(1,11.0) (2,26.0) (3,32.0) (4,40.5) (5,43.0)};
  \addplot[mark=triangle*, mark size=1.7pt, line width=0.65pt, dotted, color=SoftGreen, opacity=0.70] coordinates {(1,27.5) (2,19.5) (3,46.5) (4,68.0) (5,71.0)};
  \addplot[mark=diamond*, mark size=1.7pt, line width=0.65pt, dotted, color=BurntOrange, opacity=0.65] coordinates {(1,34.0) (2,25.5) (3,38.0) (4,44.0) (5,49.0)};
  \addplot[mark=pentagon*, mark size=1.7pt, line width=0.65pt, dotted, color=RoyalPurple, opacity=0.62] coordinates {(1,6.5) (2,11.5) (3,10.5) (4,18.5) (5,22.5)};

  \nextgroupplot[ylabel={Tok./sample}, ymin=0, ymax=930]
  \addplot[mark=*, mark size=1.5pt, line width=0.65pt, dotted, color=SoftBlack, opacity=0.70] coordinates {(1,208.0) (2,355.3) (3,266.1) (4,305.6) (5,325.8)};
  \addplot[mark=square*, mark size=1.5pt, line width=0.65pt, dotted, color=SoftBlue, opacity=0.65] coordinates {(1,425.1) (2,842.8) (3,606.4) (4,657.9) (5,700.3)};
  \addplot[mark=triangle*, mark size=1.7pt, line width=0.65pt, dotted, color=SoftGreen, opacity=0.70] coordinates {(1,200.6) (2,202.8) (3,275.6) (4,360.0) (5,358.4)};
  \addplot[mark=diamond*, mark size=1.7pt, line width=0.65pt, dotted, color=BurntOrange, opacity=0.65] coordinates {(1,220.0) (2,233.8) (3,234.8) (4,274.7) (5,289.0)};
  \addplot[mark=pentagon*, mark size=1.7pt, line width=0.65pt, dotted, color=RoyalPurple, opacity=0.62] coordinates {(1,305.6) (2,330.7) (3,340.8) (4,483.2) (5,523.0)};

  \end{groupplot}
  \end{tikzpicture} 
    \vspace{-0.6ex}
    
    \textbf{C. Filter capture rates across models}\\[-0.6ex]

  {\scriptsize
  \renewcommand{\arraystretch}{1.00}
  \begin{tabular*}{\linewidth}{@{\extracolsep{\fill}}l@{}c@{}c@{}c@{}c@{}c@{}}
  \toprule
   & Q1.5 & Q7 & Q14 & L8 & G12 \\
  \midrule
  Q1.5 & 74.7 & \rcell{29}{19.7} & \rcell{37}{6.2}  & \rcell{32}{14.4} & \rcell{32}{15.0} \\
  Q7   & \rcell{28}{23.1} & 74.2 & \rcell{21}{34.5} & \rcell{31}{16.7} & \rcell{27}{24.6} \\
  Q14  & \rcell{29}{23.1} & \rcell{34}{11.8} & 84.8 & \rcell{31}{19.7} & \rcell{31}{19.3} \\
  L8   & \rcell{11}{35.7} & \rcell{24}{19.1} & \rcell{3}{45.5}  & 49.2 & \rcell{18}{26.7} \\
  G12  & \rcell{17}{36.8} & \rcell{12}{44.9} & \rcell{17}{36.6} & \rcell{32}{12.9} & 63.6 \\
  \bottomrule
  \end{tabular*}
  }
  
    \end{minipage}

    \caption{Summarized results for prefix filters over the MLIR domain.
    Panel A shows the descriptions of the two most active filters for each model; 
    parentheses report the number of triggers (i.e., the number of errors caught by the filters) 
    and the percentage of triggers 
    that each filter contributed to the entire model (e.g., the first filter for Qwen-1.5B 
    was responsible for catching 20.8\% of errors made by the LLM).
    Panel B plots compile rate and consumed tokens per sample.
    Panel C measures filter transferability: 
    each entry reports the percentage of bad unconstrained
    samples from the column model that are caught by at least one of the filters learned from the row model. 
    Diagonal entries show the rate of bad samples matched by the model's own learned filters; 
    off-diagonal entries show cross-model transfer. 
    Darker red indicates weaker transfer relative to the row model’s own coverage. 
    }
    \label{fig:mlir-constraints}
    }
  \end{figure*}

\Cref{fig:mlir-constraints} summarizes the results for the MLIR domain, with 
Panel A on the left describing the most active filters per model, and Panel B comparing 
compile rates and tokens / sample for each mode of generation.
The high trigger rates for the top filters support our hypothesis that errors made by LLMs 
are mostly clustered: e.g., for Qwen-7B, the following top filter was responsible for 35.0\% of all
filter trigger events:
\begin{lstlisting}[style=pyinline]
def filter_unknown_torch_aten_softmax_family_op(text: str) -> bool:
    import re
    return re.search(
        r'(?m)^\s*%[\w.]+\s*=\s*torch\.aten\.(?:softmax|log_softmax|logsumexp)\s',
        text,
    ) is not None
\end{lstlisting}
This filter is very simple, just ruling out three kinds of MLIR prefixes that contain 
\texttt{torch.aten.softmax}, \texttt{log\_softmax}, or \texttt{logsumexp}, all three of which 
are unsupported (undefined) MLIR operators. 
However, Qwen-7B generates programs containing these operators exceedingly often, as shown by the 
high trigger rate of this filter.
Other models show similar clustered behavior around a few error modes: on average, 
the top 3 filters for each model contributed to 48.6\% of trigger events, while the top 5 filters 
contributed to 64.2\%, from an average of 27 total filters learned 
(ranging from 20 for Qwen-7B to 32 for Qwen-14B).

In MLIR, the learned filters lead to a significant increase in the compile rate of 
generated samples, with the most significant increase observed in Qwen-14B: from 27.5\% 
unconstrained to 71.0\% after three rounds of filter learning.
Grammar-constrained decoding performs poorly for MLIR, as grammar-constrained decoding fails to 
prevent the most prominent failure modes of the model (e.g., using undefined \texttt{torch.aten} 
operators).
The increased compile rate does come at a cost of requiring more tokens per 
sample to generate due to CARS, plotted at the bottom of Panel B, with 60.5\% 
more tokens consumed on average compared to unconstrained to generate a single sample.

Panel C in \Cref{fig:mlir-constraints} investigates \emph{filter transferability} across 
different models, where the filters of a row are applied against the unconstrained 
samples of the column models (e.g., 74.7\% of Qwen-1.5B's bad unconstrained samples 
are caught by its own prefix filters).
The table shows that filter transferability is generally poor.
In other words, 
\emph{error modes differ} quite severely between models, even between models from the 
same family but of different size---for example, filters from Qwen-14B filters match 84.8\% of 
its own bad samples, but cover only 23.1\% and 11.8\% of bad samples from 
Qwen-1.5B and 7B.
The outlier is filters from Llama-8B catching bad samples from Qwen-14B relatively well---this is because 
Llama also contains a filter that catches arity issues in \texttt{torch.aten} operators, 
the most dominant error mode for Qwen-14B.

\subsection{Molecules}
\input{figs/molecules-constraints}

\Cref{fig:molecules-constraints} summarizes the results for the molecules domain, 
where we learn filters for both chemical validity and QED score.
Results for molecules display a similar trend to those for MLIR, except that some models 
(Qwen 14B in particular) are already highly competent and do not benefit significantly 
from the prefix filters.

Bad samples generated by the unconstrained LLMs are similarly centered around 
a few significant error modes, with the top 3 filters contributing to 63.9\% of all filter triggers and the 
top 5 contributing to 78.0\%.
The most active filters are generally QED filters targeting molecules with
low QED scores, partly due to the fact that 
unconstrained Qwen-7B/14B and Gemma 12B are already highly likely (> 83.5\%) to 
generate valid molecules.
For these models, the prefix filters did not increase the validity rate statistically, 
while on Qwen-1.5B and Llama-8B (which have lower unconstrained validity rates) 
the filters did succeed in increasing validity.\footnote{
  The increase for Qwen-1.5B is insignificant at 95\% bootstrap confidence intervals, 
  and significant at 90\%.
}

On the other hand, filters helped increase QED score with the exception of Qwen-1.5B and 14B.
For Qwen-1.5B, mean QED increased by 0.034 and was statistically insignificant; the main gains for 
Qwen-1.5B are in validity rates, which do exhibit a significant increase.
For Qwen-14B, mean QED only increased by 0.008 and was statistically insignificant.
Unconstrained Qwen-14B already generates high-QED molecules averaging a score of 0.745 (on a scale of 0 to 1) 
with most samples being centered around the average; this makes it hard to learn prefix filters that clearly 
split the distribution into high- and low-QED molecules, and the learned filters were only able to 
prune a small number of molecules.
On the other hand, tokens-per-sample overhead remained lower for molecules at an average increase of 31.3\% 
(9.2 more tokens).

Like MLIR, filters learned for a specific model do not generally transfer well to other models.
The one outlier is Qwen-14B's filters on Llama-8B: here, the filter 
``polycarbonyl-rich scaffolds'' for Qwen has significant overlap with the top two filters for Llama in 
Panel A.
The two Llama filters are unable to capture Qwen's error distribution, however: 
the Qwen polycarbonyl filter is a simple one-liner  \texttt{smiles.count("C(=O)") >= 3}
that triggers on a wide range of molecules, while the Llama filters are more specific.

\subsection{Natural language summaries}
 \begin{figure*}[t!]
    {\centering
    \small
    \setlength{\tabcolsep}{4pt}
    \renewcommand{\arraystretch}{1.08}

    \begin{minipage}[t]{0.52\textwidth}
    \textbf{A. Most-triggered HR prefix filters per-model}\\[-0.5ex]

  \begin{tabular}{@{}p{0.13\linewidth}p{0.82\linewidth}}
  \toprule
  Model & Top triggered filters {\scriptsize\textcolor{black!55}{[learned round]}} \\
  \midrule
  Qwen 1.5B &
    \triggerbar{Leak patterns: ``name is'', ``named'', ``identified as''}{r0}{20.9}{508}{34.7}
  \\[-1ex]
  &
    \triggerbar{Leak patterns: ``named'', ``called'', ``referred to as''}{r0}{8.6}{176}{12.0}
  \\[-0.8ex]
    \multicolumn{2}{@{}l}{\textcolor{black!40}{\makebox[\linewidth][l]{\dotfill}}}\\[-0.4ex]
    
  Qwen 7B &
    \triggerbar{Direct name leaks after identity/naming cues}{r0}{19.7}{804}{23.4}
  \\[-1ex]
  &
    \triggerbar{Role-to-name leaks (e.g. ``manager John'')}{r0}{5.9}{774}{22.5}
  \\[-0.8ex]
  \multicolumn{2}{@{}l}{\textcolor{black!40}{\makebox[\linewidth][l]{\dotfill}}}\\[-0.4ex]
  
  Qwen 14B &
    \triggerbar{Sentence-head name leaks (e.g. ``John says ...'')}{r0}{4.7}{238}{32.6}
  \\[-1ex]
  &
    \triggerbar{Direct name leaks after naming cues}{r0}{5.8}{129}{17.7}
  \\[-0.8ex]
  \multicolumn{2}{@{}l}{\textcolor{black!40}{\makebox[\linewidth][l]{\dotfill}}}\\[-0.4ex]
  
  Llama 8B &
    \triggerbar{Attribution leaks (e.g. ``approved by John'')}{r0}{3.8}{182}{24.9}
  \\[-1ex]
  &
    \triggerbar{Role leaks (e.g., ``the employee John'')}{r0}{7.9}{167}{22.8}
  \\[-0.8ex]
  \multicolumn{2}{@{}l}{\textcolor{black!40}{\makebox[\linewidth][l]{\dotfill}}}\\[-0.4ex]
  
  Gemma 12B &
    \triggerbar{Contact/coordinator/witness names}{r0}{1.5}{27}{55.1}
  \\[-1ex]
  &
    \triggerbar{Training/contact coordinator names}{r1}{0.5}{8}{16.3}
  \\
  \bottomrule
  \end{tabular}
    \end{minipage}%
    \hfill
    \begin{minipage}[t]{0.44\textwidth}
 \textbf{B. HR performance by method}\\[-1.3ex]

  {\scriptsize
  \setlength{\tabcolsep}{2.4pt}
  \begin{tabular}{@{}ccc@{}}
  \raisebox{0.35ex}{\tikz{\draw[SoftBlack, dotted, line width=0.7pt, opacity=0.70] (0,0) -- (0.35,0); \fill[SoftBlack, opacity=0.70] (0.175,0) circle (1.2pt);}} Qwen 1.5B &
  \raisebox{0.35ex}{\tikz{\draw[SoftBlue, dotted, line width=0.7pt, opacity=0.65] (0,0) -- (0.35,0); \node[draw=SoftBlue, fill=SoftBlue, opacity=0.65, inner sep=1.1pt, regular
  polygon, regular polygon sides=4] at (0.175,0) {};}} Qwen 7B &
  \raisebox{0.35ex}{\tikz{\draw[SoftGreen, dotted, line width=0.7pt, opacity=0.70] (0,0) -- (0.35,0); \node[regular polygon, regular polygon sides=3, fill=SoftGreen, opacity=0.70,
  inner sep=1.2pt] at (0.175,0) {};}} Qwen 14B \\[-0.2ex]
  \raisebox{0.35ex}{\tikz{\draw[BurntOrange, dotted, line width=0.7pt, opacity=0.65] (0,0) -- (0.35,0); \node[diamond, fill=BurntOrange, opacity=0.65, inner sep=1.2pt] at (0.175,0)
  {};}} Llama 8B &
  \raisebox{0.35ex}{\tikz{\draw[RoyalPurple, dotted, line width=0.7pt, opacity=0.62] (0,0) -- (0.35,0); \node[regular polygon, regular polygon sides=5, fill=RoyalPurple,
  opacity=0.62, inner sep=1.1pt] at (0.175,0) {};}} Gemma 12B &
  \end{tabular}
  }\\[-0.8ex]

  \begin{tikzpicture}
  \begin{groupplot}[
    group style={
      group size=1 by 2,
      vertical sep=0.6cm,
      x descriptions at=edge bottom,
    },
    width=0.96\linewidth,
    height=0.42\linewidth,
    xtick={1,2,3,4},
    xticklabels={U,R1,R2,R3},
    xticklabel style={font=\tiny},
    yticklabel style={font=\tiny},
    label style={font=\scriptsize},
    ylabel style={font=\scriptsize, yshift=-0.15em},
    grid=both,
    major grid style={black!6},
    minor grid style={black!3},
    tick align=outside,
  ]

  \nextgroupplot[
    ylabel={Safe (\%)},
    ymin=70,
    ymax=100,
    ytick={70,80,90,100},
    xticklabels={},
  ]
  \addplot[mark=*, mark size=1.5pt, line width=0.65pt, dotted, color=SoftBlack, opacity=0.70] coordinates {(1,78.8) (2,95.6) (3,96.0) (4,97.2)};
  \addplot[mark=square*, mark size=1.5pt, line width=0.65pt, dotted, color=SoftBlue, opacity=0.65] coordinates {(1,75.4) (2,89.6) (3,92.6) (4,94.2)};
  \addplot[mark=triangle*, mark size=1.7pt, line width=0.65pt, dotted, color=SoftGreen, opacity=0.70] coordinates {(1,89.6) (2,96.0) (3,96.6) (4,96.6)};
  \addplot[mark=diamond*, mark size=1.7pt, line width=0.65pt, dotted, color=BurntOrange, opacity=0.65] coordinates {(1,87.0) (2,96.2) (3,98.0) (4,98.0)};
  \addplot[mark=pentagon*, mark size=1.7pt, line width=0.65pt, dotted, color=RoyalPurple, opacity=0.62] coordinates {(1,98.2) (2,98.4) (3,98.8) (4,98.8)};

  \nextgroupplot[
    ylabel={Info count},
    ymin=0.3,
    ymax=0.54,
    ytick={0.3,0.4,0.5},
  ]
  \addplot[mark=*, mark size=1.5pt, line width=0.65pt, dotted, color=SoftBlack, opacity=0.70] coordinates {(1,0.419) (2,0.411) (3,0.407) (4,0.407)};
  \addplot[mark=square*, mark size=1.5pt, line width=0.65pt, dotted, color=SoftBlue, opacity=0.65] coordinates {(1,0.476) (2,0.468) (3,0.470) (4,0.472)};
  \addplot[mark=triangle*, mark size=1.7pt, line width=0.65pt, dotted, color=SoftGreen, opacity=0.70] coordinates {(1,0.484) (2,0.489) (3,0.488) (4,0.488)};
  \addplot[mark=diamond*, mark size=1.7pt, line width=0.65pt, dotted, color=BurntOrange, opacity=0.65] coordinates {(1,0.520) (2,0.512) (3,0.512) (4,0.513)};
  \addplot[mark=pentagon*, mark size=1.7pt, line width=0.65pt, dotted, color=RoyalPurple, opacity=0.62] coordinates {(1,0.501) (2,0.500) (3,0.500) (4,0.501)};

  \end{groupplot}
  \end{tikzpicture}

    \vspace{-0.6ex}
    \textbf{C. Filter capture rates across models}\\[-0.6ex]

  {\scriptsize
  \renewcommand{\arraystretch}{1.00}
  \begin{tabular*}{\linewidth}{@{\extracolsep{\fill}}l@{}c@{}c@{}c@{}c@{}c@{}}
  \toprule
   & Q1.5 & Q7 & Q14 & L8 & G12 \\
  \midrule
  Q1.5 & 95.3 & \rcell{8}{85.4} & \rcell{12}{80.8} & \rcell{6}{87.7} & \rcell{40}{44.4} \\
  Q7   & \rcell{6}{85.8} & 92.7 & \rcell{9}{82.7} & \rcell{2}{90.8} & \rcell{40}{55.6} \\
  Q14  & \rcell{13}{69.8} & \rcell{1}{81.3} & 82.7 & \rcell{1}{81.5} & \rcell{26}{55.6} \\
  L8   & \rcell{2}{87.7} & \rcell{6}{77.2} & \rcell{10}{67.3} & 90.8 & \rcell{40}{33.3} \\
  G12  & \gcell{3}{59.4} & \rcell{5}{48.8} & \rcell{8}{44.2} & \rcell{9}{43.1} & 55.6 \\
  \bottomrule
  \end{tabular*}
  }

    \end{minipage}

    \caption{Summarized results for prefix filters over the HR-MultiWOZ domain.
    Panels report the same information as in \Cref{fig:mlir-constraints}, except that the 
    second graph in Panel B plots displays the average number of leaks per sample instead.}
    \label{fig:hr-constraints}
    }
  \end{figure*}

\Cref{fig:hr-constraints} summarizes the experimental results for 
the natural-language summary benchmark, which again shows the same overall trends 
as MLIR and molecule generation.
For this domain, we learn prefix filters that identify information leaks in the generated summaries.
A summary is labeled safe if it does not contain any leaks; the prefix filters 
result in a significant increase in the ratio of safe summaries, with the exception of 
Gemma3-12B, which already generates safe summaries 98.2\% of the time.

Panel B of \Cref{fig:hr-constraints} shows that the information count 
(the number of unprotected data values from the HR-MultiWoz dataset correctly included in the summary)
per summary remained statistically stable as filters were added.
In other words: the learned prefix filters successfully target just information leaks, without 
being overly aggressive and preventing useful information from appearing in the summaries.

Tokens per sample overhead remained relatively low at an average increase of 26.5\% (42.7 more tokens)
across all models.
Qwen-7B however was an outlier in token efficiency, incurring an overhead of 73.8\% 
(105.6 more tokens per sample).
This difference was due to the fact that Qwen-7B repeatedly tried to generate name-leaking summaries 
for a few specific tasks in the HR-MultiWOZ dataset; the biggest offender took 2139 tokens / sample to 
generate (compared to 143.1 tokens / sample unconstrained) and contributed 37.3 of the entire token count.
In other words, the high token overhead for Qwen-7B is caused by a few pathological samples, and is 
not an overall trend.

Most failures in the natural language domain were concentrated around name leaks:
only 36.8\% of samples in unconstrained Qwen-1.5B with a leak contained a non-name leak; 
this metric was less than 25\% for all other models.
This error distribution is reflected in the top active filters in \Cref{fig:hr-constraints}, 
all of which are name-leak filters; an example filter here is:
\begin{lstlisting}[style=pyinline]
def filter_name_named_or_identified_as(seq_prefix: str) -> bool:
    return bool(re.search(
        r'\b(?:named|called|identified as|referred to as)\s+'
        r'[A-Z][a-z]+(?:\s+[A-Z][a-z]+){0,2}\b',
        seq_prefix,
    ))
\end{lstlisting}
This filter simply identifies 4 common lead-ins to name leaks followed by a capitalized word 
(e.g., \texttt{called John}).
Because all models share a similar failure mode, the summary domain is one where 
learned filters generalize relatively well across all models.

\subsection{TypeScript}
  \begin{figure*}[t!]
    {\centering
    \small
    \setlength{\tabcolsep}{4pt}
    \renewcommand{\arraystretch}{1.08}

    \begin{minipage}[t]{0.52\textwidth}
    \textbf{A. Most-triggered TypeScript prefix filters per-model}\\[-0.5ex]

    \begin{tabular}{@{}p{0.13\linewidth}p{0.82\linewidth}}
    \toprule
    Model & Top triggered filters {\scriptsize\textcolor{black!55}{[learned round]}} \\
    \midrule
  Qwen 1.5B &
    \triggerbar{In-body ES \texttt{import} declarations}{r0}{24.5}{543}{15.8}
  \\[-1ex]
  &
    \triggerbar{Stray \texttt{function} declaration fragments}{r0}{22.1}{491}{14.3}
  \\[-0.8ex]
  \multicolumn{2}{@{}l}{\textcolor{black!40}{\makebox[\linewidth][l]{\dotfill}}}\\[-0.4ex]

  Qwen 7B &
    \triggerbar{Impossible \texttt{if} conditions with adjacent identifiers}{r1}{7.6}{168}{41.1}
  \\[-1ex]
  &
    \triggerbar{Reassignment of \texttt{const} bindings}{r0}{5.3}{117}{28.6}
  \\[-0.8ex]
  \multicolumn{2}{@{}l}{\textcolor{black!40}{\makebox[\linewidth][l]{\dotfill}}}\\[-0.4ex]

  Qwen 14B &
    \triggerbar{Reassignment of \texttt{const} bindings}{r0}{29.0}{645}{77.4}
  \\[-1ex]
  &
    \triggerbar{Block-scoped variables used after block close}{r2}{5.0}{111}{13.3}
  \\[-0.8ex]
  \multicolumn{2}{@{}l}{\textcolor{black!40}{\makebox[\linewidth][l]{\dotfill}}}\\[-0.4ex]

  Llama 8B &
    \triggerbar{Use of \texttt{this} in standalone function bodies}{r1}{35.5}{789}{21.9}
  \\[-1ex]
  &
    \triggerbar{Trailing stray \texttt{function} keyword}{r0}{32.1}{714}{19.8}
  \\[-0.8ex]
  \multicolumn{2}{@{}l}{\textcolor{black!40}{\makebox[\linewidth][l]{\dotfill}}}\\[-0.4ex]

  Gemma 12B &
    \triggerbar{Unfinished branch ladders / control-flow cutoff}{r0}{0.05}{1}{33.3}
  \\[-1ex]
  &
    \triggerbar{Late comment or call-stub cutoffs}{r2}{0.05}{1}{33.3}
  \\
    \bottomrule
    \end{tabular}
    \end{minipage}%
    \hfill
    \begin{minipage}[t]{0.44\textwidth}
    \textbf{B. TypeScript performance by method}\\[-1.3ex]
    
  {\scriptsize
  \begin{tabular}{ccc}
  \raisebox{0.35ex}{\tikz{\draw[SoftBlack, dotted, line width=0.7pt, opacity=0.70] (0,0) -- (0.35,0); \fill[SoftBlack, opacity=0.70] (0.175,0) circle (1.2pt);}} Qwen 1.5B &
  \raisebox{0.35ex}{\tikz{\draw[SoftBlue, dotted, line width=0.7pt, opacity=0.65] (0,0) -- (0.35,0); \node[draw=SoftBlue, fill=SoftBlue, opacity=0.65, inner sep=1.1pt, regular
  polygon, regular polygon sides=4] at (0.175,0) {};}} Qwen 7B &
  \raisebox{0.35ex}{\tikz{\draw[SoftGreen, dotted, line width=0.7pt, opacity=0.70] (0,0) -- (0.35,0); \node[regular polygon, regular polygon sides=3, fill=SoftGreen, opacity=0.70,
  inner sep=1.2pt] at (0.175,0) {};}} Qwen 14B \\[-0.2ex]
  \raisebox{0.35ex}{\tikz{\draw[BurntOrange, dotted, line width=0.7pt, opacity=0.65] (0,0) -- (0.35,0); \node[diamond, fill=BurntOrange, opacity=0.65, inner sep=1.2pt] at (0.175,0)
  {};}} Llama 8B &
  \raisebox{0.35ex}{\tikz{\draw[RoyalPurple, dotted, line width=0.7pt, opacity=0.62] (0,0) -- (0.35,0); \node[regular polygon, regular polygon sides=5, fill=RoyalPurple,
  opacity=0.62, inner sep=1.1pt] at (0.175,0) {};}} Gemma 12B &
  \end{tabular}
  }\\[-0.8ex]
  \begin{tikzpicture}
  \begin{groupplot}[
    group style={
      group size=1 by 2,
      vertical sep=0.6cm,
      x descriptions at=edge bottom,
    },
    width=0.96\linewidth,
    height=0.42\linewidth,
    xtick={1,2,3,4},
    xticklabels={U,R1,R2,R3},
    xticklabel style={font=\tiny},
    yticklabel style={font=\tiny},
    label style={font=\scriptsize},
    ylabel style={font=\scriptsize, yshift=-0.15em},
    grid=both,
    major grid style={black!6},
    minor grid style={black!3},
    tick align=outside,
  ]

  \nextgroupplot[
    ylabel={Compile (\%)},
    ymin=0,
    ymax=100,
    xticklabels={},
  ]
  \addplot[mark=*, mark size=1.5pt, line width=0.65pt, dotted, color=SoftBlack, opacity=0.70] coordinates {(1,16.0) (2,51.6) (3,61.4) (4,70.0)};
  \addplot[mark=square*, mark size=1.5pt, line width=0.65pt, dotted, color=SoftBlue, opacity=0.65] coordinates {(1,77.0) (2,89.8) (3,88.2) (4,87.4)};
  \addplot[mark=triangle*, mark size=1.7pt, line width=0.65pt, dotted, color=SoftGreen, opacity=0.70] coordinates {(1,92.2) (2,92.8) (3,93.8) (4,93.0)};
  \addplot[mark=diamond*, mark size=1.7pt, line width=0.65pt, dotted, color=BurntOrange, opacity=0.65] coordinates {(1,45.4) (2,65.4) (3,65.4) (4,66.6)};
  \addplot[mark=pentagon*, mark size=1.7pt, line width=0.65pt, dotted, color=RoyalPurple, opacity=0.62] coordinates {(1,95.0) (2,95.0) (3,95.0) (4,95.2)};

  \nextgroupplot[
    ylabel={Pass@10 tasks},
    ymin=0,
    ymax=50,
    ytick={0,10,20,30,40,50},
  ]
  \addplot[mark=*, mark size=1.5pt, line width=0.65pt, dotted, color=SoftBlack, opacity=0.70] coordinates {(1,13) (2,25) (3,26) (4,29)};
  \addplot[mark=square*, mark size=1.5pt, line width=0.65pt, dotted, color=SoftBlue, opacity=0.65] coordinates {(1,32) (2,34) (3,34) (4,33)};
  \addplot[mark=triangle*, mark size=1.7pt, line width=0.65pt, dotted, color=SoftGreen, opacity=0.70] coordinates {(1,39) (2,40) (3,40) (4,39)};
  \addplot[mark=diamond*, mark size=1.7pt, line width=0.65pt, dotted, color=BurntOrange, opacity=0.65] coordinates {(1,30) (2,32) (3,33) (4,31)};
  \addplot[mark=pentagon*, mark size=1.7pt, line width=0.65pt, dotted, color=RoyalPurple, opacity=0.62] coordinates {(1,34) (2,34) (3,34) (4,34)};

  \end{groupplot}
  \end{tikzpicture}
    
  \vspace{-0.6ex}
  \textbf{C. Filter capture rates across models}\\[-0.6ex]

  {\scriptsize
  \renewcommand{\arraystretch}{1.00}
  \begin{tabular*}{\linewidth}{@{\extracolsep{\fill}}l@{}c@{}c@{}c@{}c@{}c@{}}
  \toprule
   & Q1.5 & Q7 & Q14 & L8 & G12 \\
  \midrule
  Q1.5 & 99.3 & \rcell{7}{82.6} & \rcell{21}{46.2} & \rcell{3}{92.3} & \gcell{1}{100.0} \\
  Q7   & \gcell{4}{79.0} & 71.3 & \gcell{1}{71.8} & \rcell{0}{70.7} & \rcell{15}{44.0} \\
  Q14  & \rcell{11}{31.2} & \rcell{22}{20.0} & 43.6 & \gcell{8}{52.7} & \rcell{22}{20.0} \\
  L8   & \gcell{1}{96.9} & \rcell{14}{61.7} & \rcell{4}{84.6} & 94.1 & \rcell{20}{48.0} \\
  G12  & \gcell{40}{71.4} & \gcell{40}{40.9} & \gcell{40}{51.3} & \gcell{40}{53.1} & 4.0 \\
  \bottomrule
  \end{tabular*}
  }
  
    \end{minipage}

    \caption{Summarized results for prefix filters over the TypeScript domain.
    Panels report the same information as in \Cref{fig:mlir-constraints}, except that 
    Panel B shows compile rates over all samples (top) and number of testing set 
    benchmarks solved at pass@10 (bottom, 50 tasks total).
    }
    \label{fig:typescript-constraints}
    }
  \end{figure*}

\Cref{fig:typescript-constraints} summarizes the results of the TypeScript domain, 
where we learn prefix filters that block programs containing errors that would prevent compilation.
For the TypeScript domain, we reiterate that filters are learned from generating 5 samples each 
for 100 of the MultiPL-E TypeScript problems, 
then generating 10 samples each for a disjoint set of 50 MultiPL-E problems.

The filters succeed in significantly increasing compile rate, as evidenced on the top graph in 
\Cref{fig:typescript-constraints}, with the exception of Gemma-12B which already
generates compiling samples with 95.0\% accuracy.
However, for most models, the increase in compile rate did not translate directly to an 
increase tasks solved at pass@10, with the exception of Qwen-1.5B: Qwen-1.5B solved 
13/50 tasks unconstrained and 29/50 tasks with filters for a 16-solved task increase, while other models
saw a maximum of two more tasks solved.
This pattern shows that while the filters are working to shift the distribution towards samples 
that are more likely to \emph{compile}, these samples are not necessarily guaranteed to be 
\emph{correct} on a specific TypeScript task---which is expected, as the filters are optimizing 
only for compile rate and not task correctness.

TypeScript is a domain where tokens per sample sometimes suffered significantly:
Panel A in \Cref{fig:typescript-constraints} shows that the most active filters for Qwen-1.5B and 
Llama-8B were triggered over 1000 times.
Overall, Qwen-1.5B went from requiring 46.1 tokens per sample unconstrained to 
467.0 tokens per sample with all filters; Llama 8B went from 167.7 tokens per sample unconstrained 
to 838.8 tokens per sample with all filters (914.1\% / 400.2\% relative overhead respectively).
Overhead was more modest for other models, at around 
80\% for the remaining Qwen models and 
0\% for Gemma - this is because Gemma had too few invalid examples to learn filters from, 
and the few filters that were learned barely fired.

Filters for TypeScript generalize better than those for MLIR or molecules, as illustrated in 
Panel C of \Cref{fig:typescript-constraints}.
However, the TypeScript filters also show how filters from one model can be 
\emph{distributionally unsound} (i.e., filter out even valid samples too often) on other models: 
for example, consider the fact that the Qwen-1.5B filters have a 100.0\% fire rate on invalid samples from 
Gemma-12B.
The invalid samples in Gemma are actually all caught by a single Qwen-1.5B filter:
\begin{lstlisting}[style=pyinline]
def detect_unnecessary_whitespace(seq_prefix: str) -> bool:
    s = seq_prefix.strip()
    return True if (not s) else False
\end{lstlisting}
This filter detects whether the prefix is pure whitespace or padding, and rejects the current sample if so.
In general, this filter is \emph{unsound}: TypeScript functions can be compiled even if they start with a large 
number of spaces or newlines.
However, this filter is \emph{distributionally sound} for Qwen-1.5B: in our learning samples, 
all samples that start with whitespace eventually become invalid (e.g., by not generating a \texttt{return} 
statement for a function that returns a value), while 
all valid samples start with non-whitespace tokens.
On the other hand, Gemma does not display this particular mode of behavior---it often generates 
valid samples that start with newlines.
Thus this filter is distributionally unsound on Gemma, and ends up filtering many valid samples as well.
The Qwen-1.5B-Gemma capture rate after removing this filter is still 56.0\%., showing that TypeScript 
still does generalize much better than molecules or MLIR: most errors from LLMs are centered around the 
error codes 
TS1005 (syntax error---missing punctuation such as \texttt{;}), 
TS2355 (function not returning a return value), 
and TS2322 (assignment to a variable of incompatible type), which share filters.

\subsection{SQL}
  \begin{figure*}[t!]
    {\centering
    \small
    \setlength{\tabcolsep}{4pt}
    \renewcommand{\arraystretch}{1.08}

    \begin{minipage}[t]{0.52\textwidth}
    \textbf{A. Most-triggered SQL prefix filters per-model}\\[-0.5ex]

    \begin{tabular}{@{}p{0.13\linewidth}p{0.82\linewidth}}
    \toprule
    Model & Top triggered filters {\scriptsize\textcolor{black!55}{[learned round]}} \\
    \midrule
    Qwen 1.5B &
      \triggerbar{\texttt{AVG(...)} over nonnumeric schema columns}{r1}{5.1}{182}{15.8}
    \\[-1ex]
    &
      \triggerbar{Aliased table later reused by original table name}{r0}{3.8}{135}{11.7}
    \\[-0.8ex]
    \multicolumn{2}{@{}l}{\textcolor{black!40}{\makebox[\linewidth][l]{\dotfill}}}\\[-0.4ex]

    Qwen 7B &
      \triggerbar{Nonexistent columns on \texttt{results}/\texttt{laptimes}}{r1}{18.7}{45}{58.4}
    \\[-1ex]
    &
      \triggerbar{Formula-1 time strings cast to numeric types}{r1}{6.7}{16}{20.8}
    \\[-0.8ex]
    \multicolumn{2}{@{}l}{\textcolor{black!40}{\makebox[\linewidth][l]{\dotfill}}}\\[-0.4ex]

    Qwen 14B &
      \triggerbar{Mixed aggregate/raw \texttt{SELECT} without \texttt{GROUP BY}}{r0}{21.1}{249}{65.9}
    \\[-1ex]
    &
      \triggerbar{Scalar \texttt{races.name} subqueries without disambiguation}{r1}{5.3}{63}{16.7}
    \\[-0.8ex]
    \multicolumn{2}{@{}l}{\textcolor{black!40}{\makebox[\linewidth][l]{\dotfill}}}\\[-0.4ex]

    Llama 8B &
      \triggerbar{Duplicate top-level relation aliases}{r1}{9.0}{7}{28.0}
    \\[-1ex]
    &
      \triggerbar{\texttt{LIKE} applied to non-text columns}{r0}{6.4}{5}{20.0}
    \\[-0.8ex]
    \multicolumn{2}{@{}l}{\textcolor{black!40}{\makebox[\linewidth][l]{\dotfill}}}\\[-0.4ex]

    Gemma 12B &
      \triggerbar{Numeric aggregates over text columns}{r2}{16.5}{16}{51.6}
    \\[-1ex]
    &
      \triggerbar{Numeric columns compared to nonnumeric strs}{r1}{13.4}{13}{41.9}
    \\
    \bottomrule
    \end{tabular}
    \end{minipage}%
    \hfill
    \begin{minipage}[t]{0.44\textwidth}
    \textbf{B. SQL performance by method}\\[-1.3ex]

  {\scriptsize
  \begin{tabular}{ccc}
  \raisebox{0.35ex}{\tikz{\draw[SoftBlack, dotted, line width=0.7pt, opacity=0.70] (0,0) -- (0.35,0); \fill[SoftBlack, opacity=0.70] (0.175,0) circle (1.2pt);}} Qwen 1.5B &
  \raisebox{0.35ex}{\tikz{\draw[SoftBlue, dotted, line width=0.7pt, opacity=0.65] (0,0) -- (0.35,0); \node[draw=SoftBlue, fill=SoftBlue, opacity=0.65, inner sep=1.1pt, regular
  polygon, regular polygon sides=4] at (0.175,0) {};}} Qwen 7B &
  \raisebox{0.35ex}{\tikz{\draw[SoftGreen, dotted, line width=0.7pt, opacity=0.70] (0,0) -- (0.35,0); \node[regular polygon, regular polygon sides=3, fill=SoftGreen, opacity=0.70,
  inner sep=1.2pt] at (0.175,0) {};}} Qwen 14B \\[-0.2ex]
  \raisebox{0.35ex}{\tikz{\draw[BurntOrange, dotted, line width=0.7pt, opacity=0.65] (0,0) -- (0.35,0); \node[diamond, fill=BurntOrange, opacity=0.65, inner sep=1.2pt] at (0.175,0)
  {};}} Llama 8B &
  \raisebox{0.35ex}{\tikz{\draw[RoyalPurple, dotted, line width=0.7pt, opacity=0.62] (0,0) -- (0.35,0); \node[regular polygon, regular polygon sides=5, fill=RoyalPurple,
  opacity=0.62, inner sep=1.1pt] at (0.175,0) {};}} Gemma 12B &
  \end{tabular}
  }\\[-0.8ex]

  \begin{tikzpicture}
  \begin{groupplot}[
    group style={
      group size=1 by 2,
      vertical sep=0.6cm,
      x descriptions at=edge bottom,
    },
    width=0.96\linewidth,
    height=0.42\linewidth,
    xtick={1,2,3,4,5},
    xticklabels={U,Gr,R1,R2,R3},
    xticklabel style={font=\tiny},
    yticklabel style={font=\tiny},
    label style={font=\scriptsize},
    ylabel style={font=\scriptsize, yshift=-0.15em},
    grid=both,
    major grid style={black!6},
    minor grid style={black!3},
    tick align=outside,
  ]

  \nextgroupplot[
    ylabel={Compile (\%)},
    ymin=20,
    ymax=86,
    xticklabels={},
  ]
  \addplot[mark=*, mark size=1.5pt, line width=0.65pt, dotted, color=SoftBlack, opacity=0.70] coordinates {(1,25.2) (2,23.8) (3,30.0) (4,37.0) (5,37.8)};
  \addplot[mark=square*, mark size=1.5pt, line width=0.65pt, dotted, color=SoftBlue, opacity=0.65] coordinates {(1,63.0) (2,60.5) (3,64.0) (4,67.5) (5,67.5)};
  \addplot[mark=triangle*, mark size=1.7pt, line width=0.65pt, dotted, color=SoftGreen, opacity=0.70] coordinates {(1,80.5) (2,79.0) (3,82.0) (4,82.2) (5,83.0)};
  \addplot[mark=diamond*, mark size=1.7pt, line width=0.65pt, dotted, color=BurntOrange, opacity=0.65] coordinates {(1,45.5) (2,50.2) (3,46.2) (4,47.5) (5,47.8)};
  \addplot[mark=pentagon*, mark size=1.7pt, line width=0.65pt, dotted, color=RoyalPurple, opacity=0.62] coordinates {(1,78.2) (2,78.8) (3,78.2) (4,79.5) (5,81.2)};

  \nextgroupplot[
    ylabel={Pass@10 tasks},
    ymin=0,
    ymax=40,
    ytick={0,10,20,30,40},
  ]
  \addplot[mark=*, mark size=1.5pt, line width=0.65pt, dotted, color=SoftBlack, opacity=0.70] coordinates {(1,5) (2,7) (3,4) (4,6) (5,6)};
  \addplot[mark=square*, mark size=1.5pt, line width=0.65pt, dotted, color=SoftBlue, opacity=0.65] coordinates {(1,10) (2,11) (3,10) (4,10) (5,10)};
  \addplot[mark=triangle*, mark size=1.7pt, line width=0.65pt, dotted, color=SoftGreen, opacity=0.70] coordinates {(1,13) (2,15) (3,14) (4,14) (5,14)};
  \addplot[mark=diamond*, mark size=1.7pt, line width=0.65pt, dotted, color=BurntOrange, opacity=0.65] coordinates {(1,7) (2,9) (3,8) (4,8) (5,8)};
  \addplot[mark=pentagon*, mark size=1.7pt, line width=0.65pt, dotted, color=RoyalPurple, opacity=0.62] coordinates {(1,12) (2,14) (3,12) (4,12) (5,13)};

  \end{groupplot}
  \end{tikzpicture}

  \vspace{-0.6ex}
  \textbf{C. Filter capture rates across models}\\[-0.6ex]

  {\scriptsize
  \renewcommand{\arraystretch}{1.00}
  \begin{tabular*}{\linewidth}{@{\extracolsep{\fill}}l@{}c@{}c@{}c@{}c@{}c@{}}
  \toprule
   & Q1.5 & Q7 & Q14 & L8 & G12 \\
  \midrule
  Q1.5 & 74.9 & \rcell{19}{39.2} & \rcell{15}{47.4} & \rcell{10}{56.4} & \rcell{19}{40.2} \\
  Q7   & \rcell{12}{19.1} & 27.0 & \gcell{6}{30.8} & \rcell{3}{25.2} & \rcell{21}{12.6} \\
  Q14  & \rcell{18}{25.8} & \rcell{25}{17.6} & 46.2 & \rcell{20}{23.4} & \rcell{4}{41.4} \\
  L8   & \gcell{39}{14.4} & \rcell{29}{2.0} & \rcell{19}{3.8} & 7.3 & \rcell{2}{6.9} \\
  G12  & \rcell{12}{17.1} & \rcell{37}{2.0} & \rcell{21}{11.5} & \rcell{21}{11.5} & 24.1 \\
  \bottomrule
  \end{tabular*}
  }

    \end{minipage}

    \caption{Summarized results for prefix filters over the SQL domain.
    Panels report the same information as in \Cref{fig:mlir-constraints}, except that
    Panel B shows sample-level oracle-valid rates (top) and number of testing-set
    queries solved at pass@10 (bottom, 40 tasks total).}
    \label{fig:sql-constraints}
    }
  \end{figure*}

\Cref{fig:sql-constraints} summarizes the results of the SQL domain.
SQL is a domain where compile rates do not show a statistically significant increase: 
only Qwen-1.5B, which increases from 25.2\% unconstrained to 37.8\% with all filters; 
the remaining models do exhibit a mild increase, but the increase is not statistically significant 
at the 95\% bootstrap interval.
Pass@10 rates remain similarly stable, even for Qwen-1.5B: the non-increase in pass\@10 is
arguable expected as discussed for TypeScript in \Cref{sec:results}, as we learn filters only for
oracle validity, not task completion.

Qwen-14B and Gemma-12B have around 80\% success rates unconstrained; that their compile rate 
does not significantly increase is consistent with the rest of our observations that 
prefix filters struggle on improving performance for domains that LLMs are already competent at.
Qwen-7B does show a mild increase from 63.0\% compile unconstrained to 67.5\% with all filters; 
the main problem with Qwen-7B was that a learning set size of 130 samples was too small with 
a compile rate of 63.0\%, resulting in 48 invalid samples split across the 7 error buckets: 
the model did not have enough negative examples to learn generalizable prefix filters.

The outlier is Llama-8B, which failed to produce an increase in compile rate for a different reason:
the model failed to learn prefix filters because many of the failure modes were non-prefix-catchable.
For example, one of the filters proposed by \textsc{QueryLargeLLM} was (simplified for presentation):
\begin{lstlisting}[style=pyinline]
def filter_qualified_column_mismatch_on_bound_table(query_prefix: str) -> bool:
     q = query_prefix.lower()

     # Intended: catch references to non-existent columns, e.g. results.status
     if "from results" in q or "join results" in q:
         if re.search(r"\bresults\.(name|year|round|status)\b", q):
             return True

     # False-positive problem: `results.status` is bad, but `results.statusid` is valid
     return False
\end{lstlisting}

This filter is valid in principle: Llama-8B often fails by accessing the non-existent \texttt{results.status} 
column.
However, there does exists \texttt{results.statusId} column inside the database, and thus a prefix filter 
targeting \texttt{results.status} becomes unsound, as it would trigger on valid queries containing 
\texttt{results.statusId} as well.
These filters were thus rejected during the validation phase; consequently, \name could not learn 
filters that captured Llama-8B's most significant error modes, failing to improve performance.
This case illustrates a failure mode of \name discussed in \Cref{sec:discussion}: 
when error modes are not soundly capturable by prefix filters.

Filter transfer rates for SQL remain low; the models display different patterns of errors.

\section{Filter coverage statistics with intervals}
\label{app:filter-intervals}
 \newcommand{\cicell}[4]{%
    \cellcolor{cyan!#1!white}%
    \begin{tabular}{@{}c@{}}
      #2\%\\[-0.7ex]
      {\tiny [#3,#4]}
    \end{tabular}%
  }

  \newcommand{\plaincicell}[3]{%
    \begin{tabular}{@{}c@{}}
      #1\%\\[-0.7ex]
      {\tiny [#2,#3]}
    \end{tabular}%
  }

  \newcommand{\ratiocell}[3]{%
    \begin{tabular}{@{}c@{}}
      #1\\[-0.7ex]
      {\tiny [#2,#3]}
    \end{tabular}%
  }

  \begin{figure*}[t!]
  {\centering
  \small
  \setlength{\tabcolsep}{2.5pt}
  \renewcommand{\arraystretch}{1.10}

  \begin{minipage}[t]{0.60\textwidth}
  {\scriptsize
  \renewcommand{\arraystretch}{1.05}
  \begin{tabular*}{\linewidth}{@{\extracolsep{\fill}}llccccc@{}}
  \toprule
  \textbf{Domain} & & \textbf{Q1.5} & \textbf{Q7} & \textbf{Q14} & \textbf{L8} & \textbf{G12} \\
  \midrule
  MLIR & Full &
  \cicell{34}{74.7}{68.1}{80.8} &
  \cicell{33}{74.2}{67.4}{80.3} &
  \cicell{38}{84.8}{78.6}{90.3} &
  \cicell{22}{49.2}{40.9}{57.6} &
  \cicell{29}{64.2}{57.2}{71.1} \\
  (catch \%) & Top-3 &
  \cicell{15}{33.5}{26.9}{40.7} &
  \cicell{23}{51.7}{44.4}{59.0} &
  \cicell{18}{40.0}{32.4}{48.3} &
  \cicell{10}{22.0}{15.2}{28.8} &
  \cicell{12}{27.3}{21.4}{33.7} \\

  \cdashline{1-7}[0.4pt/1pt]

  Molec. & Full &
  \cicell{20}{44.9}{37.7}{52.7} &
  \cicell{26}{58.1}{48.4}{67.7} &
  \cicell{21}{46.5}{36.0}{57.0} &
  \cicell{27}{59.6}{52.4}{66.9} &
  \cicell{33}{74.3}{66.1}{82.6} \\
  (catch \%) & Top-3 &
  \cicell{8}{16.8}{11.4}{22.8} &
  \cicell{16}{35.5}{25.8}{45.2} &
  \cicell{17}{38.4}{27.9}{48.8} &
  \cicell{22}{48.2}{40.4}{56.0} &
  \cicell{28}{62.4}{53.2}{71.6} \\

  \cdashline{1-7}[0.4pt/1pt]

  Summ. & Full &
  \cicell{43}{95.3}{90.6}{99.1} &
  \cicell{42}{92.7}{87.8}{96.7} &
  \cicell{37}{82.7}{71.2}{92.3} &
  \cicell{41}{90.8}{83.1}{96.9} &
  \cicell{25}{55.6}{22.2}{88.9} \\
  (catch \%) & Top-3 &
  \cicell{34}{75.5}{67.0}{83.0} &
  \cicell{22}{48.8}{39.8}{57.7} &
  \cicell{19}{42.3}{28.8}{55.8} &
  \cicell{15}{32.3}{21.5}{43.1} &
  \cicell{15}{33.3}{0.0}{66.7} \\

  \cdashline{1-7}[0.4pt/1pt]

  TS & Full &
  \cicell{45}{99.3}{98.3}{100.0} &
  \cicell{32}{71.3}{63.5}{79.1} &
  \cicell{33}{74.4}{59.0}{87.2} &
  \cicell{42}{94.1}{91.2}{96.7} &
  \cicell{2}{4.0}{0.0}{12.0} \\
  (catch \%) & Top-3 &
  \cicell{34}{75.2}{71.0}{79.3} &
  \cicell{7}{16.5}{9.6}{23.5} &
  \cicell{16}{35.9}{20.5}{51.3} &
  \cicell{26}{57.5}{51.6}{63.4} &
  \cicell{0}{0.0}{0.0}{0.0} \\

  \cdashline{1-7}[0.4pt/1pt]

  SQL & Full &
  \cicell{34}{75.3}{70.2}{79.9} &
  \cicell{12}{27.0}{20.3}{34.5} &
  \cicell{21}{46.2}{34.6}{56.4} &
  \cicell{4}{8.7}{5.0}{12.8} &
  \cicell{11}{24.1}{16.1}{33.3} \\
  (catch \%) & Top-3 &
  \cicell{14}{32.1}{27.1}{37.5} &
  \cicell{11}{24.3}{17.6}{31.1} &
  \cicell{18}{41.0}{30.8}{52.6} &
  \cicell{2}{5.0}{2.3}{8.3} &
  \cicell{11}{24.1}{14.9}{33.3} \\
  \bottomrule
  \end{tabular*}
  }
  \end{minipage}%
  \hfill
  \begin{minipage}[t]{0.37\textwidth}

  {\scriptsize
  \renewcommand{\arraystretch}{1.12}
  \begin{tabular*}{\linewidth}{@{\extracolsep{\fill}}lccc@{}}
  \toprule
  \textbf{Domain} & \textbf{Own} & \textbf{Cross} & \textbf{Ratio} \\
  \midrule
  MLIR &
  \plaincicell{69.4}{66.4}{72.5} &
  \plaincicell{24.4}{22.2}{26.6} &
  \cellcolor{orange!28!white}\ratiocell{0.35}{0.32}{0.38} \\

  Molec. &
  \plaincicell{56.7}{52.8}{60.6} &
  \plaincicell{13.9}{12.3}{15.4} &
  \cellcolor{orange!35!white}\ratiocell{0.24}{0.22}{0.27} \\

  Summ. &
  \plaincicell{83.4}{76.3}{90.2} &
  \plaincicell{68.1}{63.1}{73.4} &
  \cellcolor{teal!22!white}\ratiocell{0.82}{0.75}{0.89} \\

  TS &
  \plaincicell{68.6}{65.0}{72.2} &
  \plaincicell{65.3}{62.5}{68.2} &
  \cellcolor{teal!31!white}\ratiocell{0.95}{0.90}{1.01} \\

  SQL &
  \plaincicell{36.3}{32.9}{39.7} &
  \plaincicell{22.8}{20.9}{24.8} &
  \cellcolor{orange!12!white}\ratiocell{0.63}{0.57}{0.70} \\
  \bottomrule
  \end{tabular*}
  }
  \end{minipage}

  \caption{
    Panels B (left) and C (right) from \Cref{fig:filter-table} annotated with
    95\% bootstrap confidence intervals, represented as the bracketed intervals on the bottom.
    Noise is centered on cases where unconstrained models are already competent and the
    number of samples available is low (e.g., Summary \& Gemma-12B, or TypeScript \& Gemma-12B).
  }
  \label{fig:filter-table-intervals}
  }
  \end{figure*}

\Cref{fig:filter-table-intervals} presents a version of panels B and C 
from \Cref{fig:filter-table} with 95\% bootstrap confidence intervals.

\newpage
\section*{NeurIPS Paper Checklist}


\begin{enumerate}

\item {\bf Claims}
    \item[] Question: Do the main claims made in the abstract and introduction accurately reflect the paper's contributions and scope?
    \item[] Answer: \answerYes{} 
    \item[] Justification: The paper's main contributions are learning error patterns via prefix filters and applying them 
      to constrained generation, which matches the content in the abstract and introduction.

\item {\bf Limitations}
    \item[] Question: Does the paper discuss the limitations of the work performed by the authors?
    \item[] Answer: \answerYes{} 
    \item[] Justification: See \Cref{sec:results} and \Cref{sec:discussion}, which discusses when 
      our approach becomes inefficient or not applicable.

\item {\bf Theory assumptions and proofs}
    \item[] Question: For each theoretical result, does the paper provide the full set of assumptions and a complete (and correct) proof?
    \item[] Answer: \answerNA{} 
    \item[] Justification: Our results are mostly empircal and we do not state or prove a formal theorem.

    \item {\bf Experimental result reproducibility}
    \item[] Question: Does the paper fully disclose all the information needed to reproduce the main experimental results of the paper to the extent that it affects the main claims and/or conclusions of the paper (regardless of whether the code and data are provided or not)?
    \item[] Answer: \answerYes{} 
    \item[] Justification: Our algorithms to learn error patterns, prefix filters, and use them for constrained generation are 
      detailed in \Cref{sec:prefix}.
      We provide an explanation of our experimental setup in \Cref{sec:experiments} and \Cref{app:setup}. 

\item {\bf Open access to data and code}
    \item[] Question: Does the paper provide open access to the data and code, with sufficient instructions to faithfully reproduce the main experimental results, as described in supplemental material?
    \item[] Answer: \answerYes{} 
    \item[] Justification: We provide our code and experimental results as supplementary material.

\item {\bf Experimental setting/details}
    \item[] Question: Does the paper specify all the training and test details (e.g., data splits, hyperparameters, how they were chosen, type of optimizer) necessary to understand the results?
    \item[] Answer: \answerYes{} 
    \item[] Justification: See \Cref{sec:experiments} and \Cref{app:setup}, which detail our experimental setup.

\item {\bf Experiment statistical significance}
    \item[] Question: Does the paper report error bars suitably and correctly defined or other appropriate information about the statistical significance of the experiments?
    \item[] Answer: \answerYes{} 
    \item[] Justification: Our experiments draw multiple samples and tasks for a single domain, 
      and comparisons made in the paper are statistically supported via the bootstraping method 
      proposed by~\citet{bootstrap}. We use a 95\% bootstrap confidence interval 
      with 10000 bootstrap samples
      unless otherwise noted, and a comparison 
      is considered statistically significant if the difference interval does not contain zero. 

\item {\bf Experiments compute resources}
    \item[] Question: For each experiment, does the paper provide sufficient information on the computer resources (type of compute workers, memory, time of execution) needed to reproduce the experiments?
    \item[] Answer: \answerYes{} 
    \item[] Justification: See \Cref{sec:experiments} and \Cref{app:hardware} for details on our hardware setup.
    
\item {\bf Code of ethics}
    \item[] Question: Does the research conducted in the paper conform, in every respect, with the NeurIPS Code of Ethics \url{https://neurips.cc/public/EthicsGuidelines}?
    \item[] Answer: \answerYes{} 
    \item[] Justification: We believe our paper does not pose any immediate societal, human, or harmful impact and follow the code of ethics.

\item {\bf Broader impacts}
    \item[] Question: Does the paper discuss both potential positive societal impacts and negative societal impacts of the work performed?
    \item[] Answer: \answerNA{} 
    \item[] Justification: We believe our paper does not pose any immediate societal impact, primarily investigating LLMs themselves 
      and ways to increase their performance, as opposed to applying LLMs towards a social problem.
    
\item {\bf Safeguards}
    \item[] Question: Does the paper describe safeguards that have been put in place for responsible release of data or models that have a high risk for misuse (e.g., pre-trained language models, image generators, or scraped datasets)?
    \item[] Answer: \answerNA{} 
    \item[] Justification: We believe our paper does not pose any immediate harmful impact, primarily investigating LLMs themselves 
      and ways to increase their performance, as opposed to applying LLMs towards a social problem.

\item {\bf Licenses for existing assets}
    \item[] Question: Are the creators or original owners of assets (e.g., code, data, models), used in the paper, properly credited and are the license and terms of use explicitly mentioned and properly respected?
    \item[] Answer: \answerYes{} 
    \item[] Justification: All assets have been credited where necessary.

\item {\bf New assets}
    \item[] Question: Are new assets introduced in the paper well documented and is the documentation provided alongside the assets?
    \item[] Answer: \answerNA{} 
    \item[] Justification: This paper does not include any new assets outside of code.

\item {\bf Crowdsourcing and research with human subjects}
    \item[] Question: For crowdsourcing experiments and research with human subjects, does the paper include the full text of instructions given to participants and screenshots, if applicable, as well as details about compensation (if any)? 
    \item[] Answer: \answerNA{} 
    \item[] Justification: The paper does not involve crowdsourcing nor research with human subjects.

\item {\bf Institutional review board (IRB) approvals or equivalent for research with human subjects}
    \item[] Question: Does the paper describe potential risks incurred by study participants, whether such risks were disclosed to the subjects, and whether Institutional Review Board (IRB) approvals (or an equivalent approval/review based on the requirements of your country or institution) were obtained?
    \item[] Answer: \answerNA{} 
    \item[] Justification: The paper does not involve crowdsourcing nor research with human subjects.

\item {\bf Declaration of LLM usage}
    \item[] Question: Does the paper describe the usage of LLMs if it is an important, original, or non-standard component of the core methods in this research? Note that if the LLM is used only for writing, editing, or formatting purposes and does \emph{not} impact the core methodology, scientific rigor, or originality of the research, declaration is not required.
    \item[] Answer: \answerNA{} 
    \item[] Justification: LLMs were used to assist writing and programming. 

\end{enumerate}

\end{document}